\documentclass[10pt,twocolumn,letterpaper]{article}

\usepackage{cvpr}      
\toggletrue{cvprfinal}

\usepackage[T1]{fontenc}
\usepackage[utf8]{inputenc}
\usepackage[warn]{textcomp}

\usepackage{siunitx}

\usepackage{enumitem} 
\usepackage{overpic} 
\usepackage{color}
\usepackage{xspace}
\usepackage{xfrac}
\usepackage{colortbl}
\usepackage{bm}

\usepackage{caption}
\usepackage{subcaption}
\usepackage{wrapfig}
\usepackage{multirow}

\usepackage{graphicx}
\usepackage{amsmath}
\usepackage{amssymb}
\usepackage{booktabs}

\usepackage{tikz}
\usepackage{pgfplots}
\pgfplotsset{compat=1.14}

\usetikzlibrary{shapes}
\usetikzlibrary{calc}
\usetikzlibrary{backgrounds}
\usetikzlibrary{positioning}
\usetikzlibrary{arrows}
\usetikzlibrary{arrows.meta}
\usetikzlibrary{decorations.text}    
\usetikzlibrary{fit}
\usetikzlibrary{spy}
\usetikzlibrary{3d}
\usepackage{tikz-3dplot}

\usepackage[skins]{tcolorbox}

\definecolor{turquoise}{cmyk}{0.65,0,0.1,0.3}
\definecolor{purple}{rgb}{0.65,0,0.65}
\definecolor{dark_green}{rgb}{0, 0.5, 0}
\definecolor{orange}{rgb}{0.8, 0.6, 0.2}
\definecolor{red}{rgb}{0.8, 0.2, 0.2}
\definecolor{darkred}{rgb}{0.6, 0.1, 0.05}
\definecolor{blueish}{rgb}{0.0, 0.3, .6}
\definecolor{light_gray}{rgb}{0.7, 0.7, .7}
\definecolor{pink}{rgb}{1, 0, 1}
\definecolor{greyblue}{rgb}{0.25, 0.25, 1}

\definecolor{bestcol}{RGB}{254,196,79}
\newcommand{\best}[1]{\cellcolor{bestcol} \textbf{#1}}
\definecolor{secondbestcol}{RGB}{255,247,188}
\newcommand{\secondbest}[1]{\cellcolor{secondbestcol} #1}

\newcommand{\vect}[1]{\bm{#1}}

\newcommand{\fig}[1]{Fig.~\ref{fig:#1}}

\newcommand{\Table}[1]{Table~\ref{tab:#1}}

\makeatletter
\DeclareRobustCommand\onedot{\futurelet\@let@token\@onedot}
\def\@onedot{\ifx\@let@token.\else.\null\fi\xspace}

\def\vs{\emph{vs}\onedot}
 
\def\etal{\emph{et al}\onedot}

\makeatother

\newcommand{\inlinesection}[1]{\vspace{0.05cm} \noindent {\bf #1}}
\newcommand{\titlecaption}[2]{\caption{\textbf{#1.}\xspace#2}}

\newcommand{\titlecaptionof}[3]{\captionof{#1}{\textbf{#2.}\xspace#3}}

\newcommand{\tightsubsection}[1]{\vspace{-0.25em}\subsection{#1}\vspace{-0.25em}}

\usepackage{pifont}
\usepackage{blindtext}

\definecolor{mpurple}{RGB}{106,27,154}
\definecolor{mpurplelight}{RGB}{206,147,216}

\definecolor{mblue}{RGB}{40,53,147}
\definecolor{mbluelight}{RGB}{159,168,218}

\definecolor{mteal}{RGB}{0,105,92}
\definecolor{mteallight}{RGB}{128,203,196}

\definecolor{morangelight}{RGB}{255,171,145}

\definecolor{mgrayblue}{RGB}{55,71,79}
\definecolor{mgraybluelight}{RGB}{176,190,197}

\definecolor{mamber}{RGB}{255,143,0}
\definecolor{mamberlight}{RGB}{255,224,130}

\definecolor{mdeeporange}{RGB}{216,67,21}
\definecolor{morange}{RGB}{245,124,0}
\definecolor{myellow}{RGB}{253,216,53}

\definecolor{mgreen}{RGB}{85,139,47}
\definecolor{mgreenlight}{RGB}{174,213,129}

\definecolor{mred}{RGB}{198,40,40}
\definecolor{mredlight}{RGB}{239,154,154}

\definecolor{corange1}{RGB}{254,237,222}
\definecolor{corange2}{RGB}{253,190,133}
\definecolor{corange3}{RGB}{253,141,60}
\definecolor{corange4}{RGB}{230,85,13}
\definecolor{corange5}{RGB}{166,54,3}

\definecolor{cblue1}{RGB}{239,243,255}
\definecolor{cblue2}{RGB}{189,215,231}
\definecolor{cblue3}{RGB}{107,174,214}
\definecolor{cblue4}{RGB}{49,130,189}
\definecolor{cblue5}{RGB}{8,81,156}

\definecolor{cgray1}{RGB}{247,247,247}
\definecolor{cgray2}{RGB}{204,204,204}
\definecolor{cgray3}{RGB}{150,150,150}
\definecolor{cgray4}{RGB}{99,99,99}
\definecolor{cgray5}{RGB}{37,37,37}

\definecolor{cred1}{RGB}{254,229,217}
\definecolor{cred2}{RGB}{252,174,145}
\definecolor{cred3}{RGB}{251,106,74}
\definecolor{cred4}{RGB}{222,45,38}
\definecolor{cred5}{RGB}{165,15,21}

\definecolor{cgreen1}{RGB}{237,248,233}
\definecolor{cgreen2}{RGB}{186,228,179}
\definecolor{cgreen3}{RGB}{116,196,118}
\definecolor{cgreen4}{RGB}{49,163,84}
\definecolor{cgreen5}{RGB}{0,109,44}

\definecolor{cdiv11}{RGB}{166,97,26}
\definecolor{cdiv12}{RGB}{223,194,125}
\definecolor{cdiv13}{RGB}{245,245,245}
\definecolor{cdiv14}{RGB}{128,205,193}
\definecolor{cdiv15}{RGB}{1,133,113}

\definecolor{cdiv21}{RGB}{208,28,139}
\definecolor{cdiv22}{RGB}{241,182,218}
\definecolor{cdiv23}{RGB}{247,247,247}
\definecolor{cdiv24}{RGB}{184,225,134}
\definecolor{cdiv25}{RGB}{77,172,38}

\definecolor{cdiv31}{RGB}{230,97,1}
\definecolor{cdiv32}{RGB}{253,184,99}
\definecolor{cdiv33}{RGB}{247,247,247}
\definecolor{cdiv34}{RGB}{178,171,210}
\definecolor{cdiv35}{RGB}{94,60,153}

\definecolor{cvprblue}{rgb}{0.21,0.49,0.74}
\usepackage[pagebackref,breaklinks,colorlinks,citecolor=cvprblue]{hyperref}

\def\paperID{*****} 
\def\confName{CVPR}
\def\confYear{2024}

\title{SF3D: Stable Fast 3D Mesh Reconstruction with UV-unwrapping and Illumination Disentanglement}
\def\ours{SF3D\xspace}

\author{Mark Boss$^1$ \quad Zixuan Huang$^{1,2\footnotemark[2]}$ \quad Aaryaman Vasishta$^1$ \quad Varun Jampani$^1$ \\ \\
$^1$Stability AI \quad $^2$UIUC\\
}

\begin{document}

\twocolumn[{%
    \renewcommand\twocolumn[1][]{#1}%
    \maketitle
    \begin{center}
    \centering
    \captionsetup{type=figure}
    \includegraphics[width=0.85\textwidth]{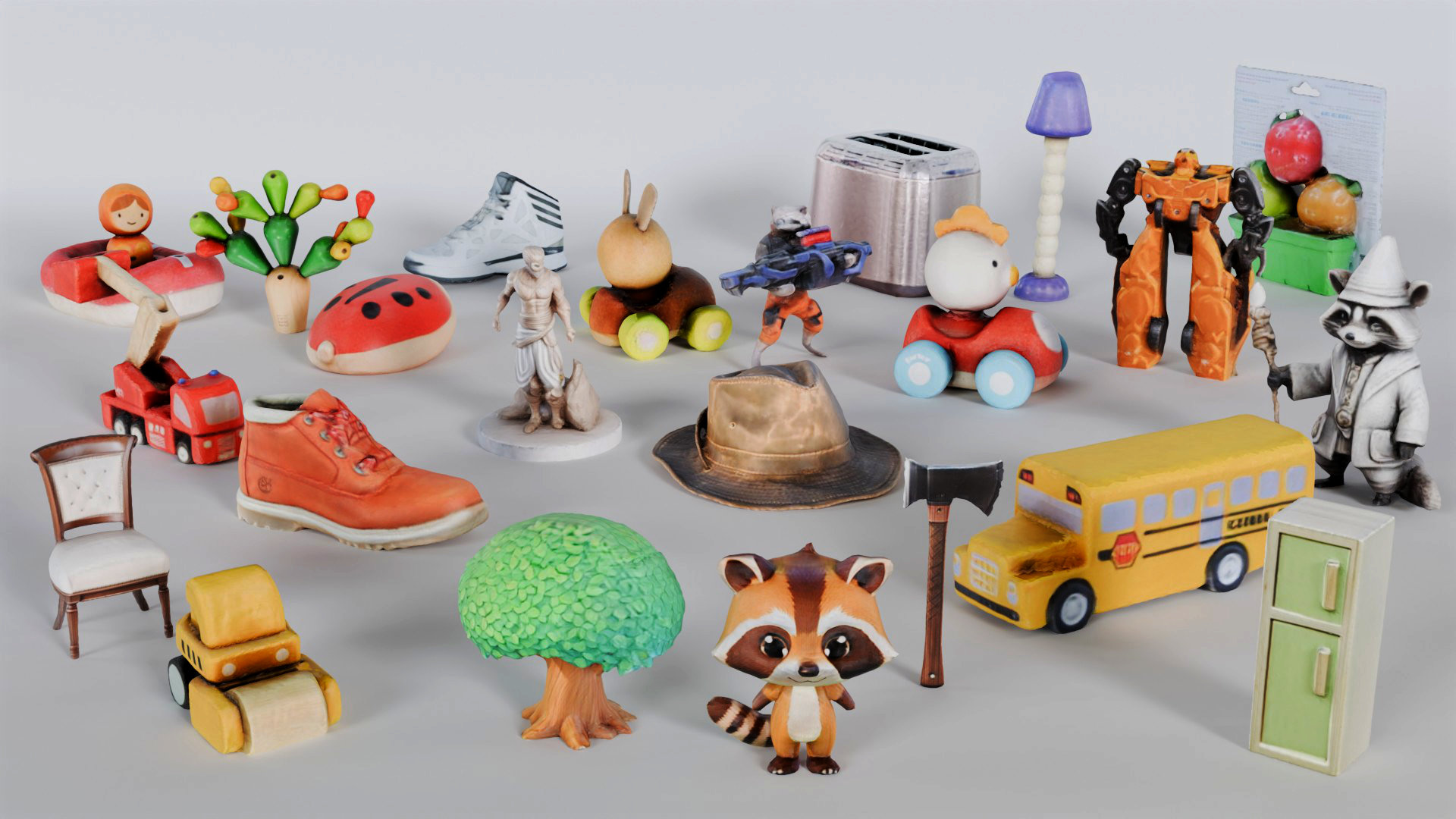}
    \caption{\textbf{SF3D Creates High-quality Object Meshes from Single Images}{ with Materials, Delighting and UV-unwrapped textured meshes in \SI{0.5}{\second}. Here, we show sample SF3D results from different input images. SF3D handles both realistic and non-realistic styles well.}}
    \label{fig:teaser}
\end{center}
}]

\renewcommand{\thefootnote}{\fnsymbol{footnote}}
\footnotetext[2]{Work done during internship at Stability AI.}

\vspace{-2mm}
\begin{abstract}
We present SF3D, a novel method for rapid and high-quality textured object mesh reconstruction from a single image in just 0.5 seconds. Unlike most existing approaches, SF3D is explicitly trained for mesh generation, incorporating a fast UV unwrapping technique that enables swift texture generation rather than relying on vertex colors. The method also learns to predict material parameters and normal maps to enhance the visual quality of the reconstructed 3D meshes. Furthermore, SF3D integrates a delighting step to effectively remove low-frequency illumination effects, ensuring that the reconstructed meshes can be easily used in novel illumination conditions. Experiments demonstrate the superior performance of SF3D over the existing techniques. 

Project page with code and model: \url{https://stable-fast-3d.github.io}
\end{abstract}
\vspace{-2mm}
\vspace{-2mm}
\section{Introduction}
\label{sec:intro}

High-quality object meshes are essential for various use cases in movies, gaming, e-commerce, and AR/VR.
In this work, we tackle the problem of generating high-quality 3D object mesh from a single image.
This is a ill-posed and challenging problem as this requires reasoning about the object's 3D shape and texture from only a single 2D projection (image) of that object.
Single-image object generation can simplify the tedious and manual object creation process.

The quality of object mesh generation from a single image has dramatically improved in the past couple of years with the advances in transformer models~\cite{hong2024lrm,openlrm,tochilkin2024triposr}, large synthetic datasets~\cite{deitke2023objaversexl} and 3D-aware image/video generative models~\cite{liu2023zero1to3,stablezero123,voleti2024sv3d,xie2024sv4d}. 
Especially, the transformer-based reconstruction models~\cite{hong2024lrm, openlrm, tochilkin2024triposr} demonstrate phenomenal generalization capabilities on real-world images despite being trained only on synthetic datasets while also generating 3D assets from a single image in under \SI{1}{\second}.

\input{fig/current_issues}

Despite this rapid progress, several issues remain in these feed-forward fast 3D reconstruction models~\cite{hong2024lrm, openlrm, tochilkin2024triposr}. These techniques often produce 3D assets that are not usable for downstream applications or require laborious manual post-processing. We identify several key issues in these techniques and propose a fast generation technique called `Stable Fast 3D' (SF3D) that generates higher quality and more usable 3D assets from single images, while also retaining the fast generation speed within 0.5 seconds on a H100 GPU. Next, we briefly introduce these issues and how we tackle those in SF3D.

\inlinesection{Light Bake-in.}
Having shadows or other illumination effects in a given input image is common. Most existing works bake these effects into textures, making the resulting 3D assets less usable. Having consistent lighting helps in easy integration into graphics pipelines. In SF3D, we propose decomposing the illumination and reflective properties by incorporating explicit illumination and a differentiable shading model using Spherical Gaussians (SG). \fig{issues} (top row) shows a sample result of SF3D, where the light bake-in is considerably reduced compared to the prior art.

\inlinesection{Vertex Coloring.}
Another issue we found in most 3D generation models is that they produce meshes with a high vertex count, using vertex coloring to represent object texture. This makes the resulting 3D assets inefficient to use in applications such as games.
A key issue is the additional processing time of UV unwrapping, which can take longer than the entire object generation. For example, xatlas~\cite{xatlas} and geogram~\cite{geogram}, can take up to \SI{30}{\second} or \SI{10}{\second} for a single asset, respectively. To tackle this, we propose a highly parallelizable fast box projection-based UV unwrapping technique to achieve a \SI{0.5}{\second} generation time. The effect of relying on vertex coloring \vs UV Unwrapping can be seen in \fig{issues} (middle row), where TripoSR captures fewer details than SF3D despite requiring $10\times$ higher polygon count.

\inlinesection{Marching Cubes Artifacts.}
The feed-forward networks often create volumetric representations such as Triplane-NeRFs~\cite{Chan2021eg3d} which are converted to meshes using the Marching Cubes (MC)~\cite{lorensen1987mc} algorithm.
MC can cause `stair-stepping' artifacts, which can be somewhat reduced by increasing the volume resolution. However, this comes at the cost of large computational overhead. In contrast, SF3D uses more efficient architecture for higher resolution triplanes and also produces meshes using DMTet~\cite{shen2021dmtet} with learned vertex displacements and normal maps, resulting in smoother mesh surfaces. \fig{issues} demonstrates the smoothness of the SF3D mesh compared to that of TripoSR.

\inlinesection{Lack of Material Properties.}
The generations from previous feed-forward techniques often look dull when they are rendered using different illuminations. This is mainly due to the lack of explicit material properties in the output generations, which can influence the light reflection. To tackle this, we predict non-spatially varying material properties. This addition is apparent when rendering different generated objects in \fig{issues} (bottom row).

With these advances, SF3D can generate high-quality 3D meshes from a single image with several desirable properties for downstream applications on both shapes (low-polygon yet smooth) and textures (Illumination disentangled UV maps with material properties).
The 3D assets are small in size (under \SI{1}{\mega\byte}) and can be created in \SI{0.5}{\second}.
For text-to-3D mesh generation, a fast Text-to-Image (T2I) model~\cite{sauer2023add} can be combined with SF3D to produce meshes in about \SI{1}{\second}.
Experimental results demonstrate higher quality results with SF3D compared to the existing works.
In short, SF3D provides a comprehensive technique for fast, high-quality 3D object generation from single images, addressing both speed and usability in practical applications.

\section{Related Work}

\inlinesection{3D Reconstruction using Image Generative Priors.}
Diffusion models~\cite{ho2020ddpm,song2020sgm} have proven to be powerful generative models for various tasks~\cite{blattmann2023align,ruiz2022dreambooth,rombach2022high, blattmann2023stable,voleti2022mcvd,Rohit2023EMU}. 
Several works such as Zero123~\cite{liu2023zero1to3} and others~\cite{deitke2023objaversexl, stablezero123, kong2024eschernet, zheng2023free3d} leverage the object priors in these diffusion models by adapting the generative models for 3D generation.
Score Distillation Sampling (SDS)~\cite{poole2022dreamfusion} is often used to optimize 3D representation using the 2D diffusion models. 
However, it was found that relying solely on the image prior does not always produce consistent multi-view results. 
This issue is improved in follow-up works~\cite{shi2023mvdream, liu2023syncdreamer, mercier2024hexagen3d, shi2023zero123pp} by simultaneously generating multiple views of an object.
Another approach is to explicitly introduce 3D awareness~\cite{ye2023consistent1to3, liu2023one2345, liu2023one2345pp}
or use a multi-view diffusion process~\cite{shi2023mvdream,weng2023consistent123,long2023wonder3d,kwak2023vivid,blattmann2023stable,melas2024im3d,voleti2024sv3d,Gao2024CAT3DCA,Zhao2024FlexiDreamerSI,Liu2023UniDreamUD} to generate 3D objects.
While diffusion models can generate videos or multi-view images relatively quickly, they require a 3D reconstruction step to create 3D mesh from a single image. 
Even with fast techniques, generating an object can still take several minutes. 
Our work focuses on fast generation speeds in \SI{0.5}{\second} for image-to-3D.

\inlinesection{Feed-Forward 3D Reconstruction.}
Recently, LRM~\cite{hong2024lrm} and the follow-up works~\cite{openlrm, tochilkin2024triposr} demonstrated that the fast and reliable 3D generation is possible using feed-forward networks trained on large synthetic datasets.
These works use large transformer models to directly predict triplanes as the volume representations, which can be ray-marched using NeRF~\cite{mildenhall2020}.
This allows these models to train on multi-view datasets, as only the rendering loss is required for training. 
A set of follow-up works address to remedy the reliance on multi-view datasets~\cite{Xie2024LRMZeroTL, Wang2023PFLRMPL, Jiang2024Real3DSU}.
This can produce meshes from images in seconds compared to the generative-prior based approaches.

Several follow-up works emerged which use Gaussian Splatting~\cite{kerbl3Dgaussians} as the representation~\cite{Szymanowicz2024SplatterIU,Zhang2024GSLRMLR,Tang2024LGMLM,Xu2024GRMLG,Zou2023TriplaneMG} or directly integrate a mesh prediction~\cite{Zhuang2024GTRIL,Xu2024InstantMeshE3,Xie2024LDMLT,Wei2024MeshLRMLR,Li2024MLRMML,Wu2024Unique3DHA}.
Another group of methods integrates diffusion models with the feed-forward models by either directly generating the triplanes~\cite{Wang2024CRMSI}, using LRM as the denoiser~\cite{Xu2023DMV3DDM}, or using it as the conditioning~\cite{Wen2024Ouroboros3DIG}.
As single-image reconstruction is a challenging task, several models leverage the extensive prior of multi-view diffusion models to generate multiple views of an object which the feed-forward model then uses to produce a 3D output~\cite{Zhang2024GSLRMLR, Zhuang2024GTRIL, Xu2024InstantMeshE3, Xie2024LDMLT, Tang2024LGMLM, Wei2024MeshLRMLR, Li2024MLRMML, Wu2024Unique3DHA, Xu2024GRMLG, Li2023Instant3DFT}.
Generally, these models learn the scene's radiance, meaning lighting information is baked into the objects.
LDM~\cite{Xie2024LDMLT} tries to remedy this by learning an additional shading color to capture the shading information. However, training requires having access to the ground-truth albedo color as a supervision signal.
Compared to LDM, our method also learns an illumination model, enabling training on regular multi-view datasets, and we predict more material properties. We also tackle fast UV unwrapping, so unlike previous works, we do not have to rely on vertex colors while keeping the 3D generation time short.

\inlinesection{Material Decomposition.}
Predicting only the radiance of an object has the downside that relighting does not produce convincing results.
Current works in single scene optimization are often based on NeRF~\cite{mildenhall2020} or Gaussian Splatting~\cite{kerbl3Dgaussians} and decompose multiple input images ($>50$) into light and materials~\cite{Boss2021, Boss2021neuralPIL, bossSAMURAIShapeMaterial2022, engelhardt2023-shinobi, zhang2021, munkberg2022nvdiffrec, hasselgrenShapeLightMaterial2022}. 
These methods often predict materials properties of a Physically Based Rendering (PBR)~\cite{Burley2012} shading model.
A few recent works tackle joint 3D shape and material generation, such as UniDream~\cite{Liu2023UniDreamUD} or Fantasia3D~\cite{chen2023fantasia3d} by optimizing the material properties using a SDS loss. 
However, this optimization takes several hours to converge.
Another set of works aims to texture existing objects~\cite{Zhang2024DreamMatHP, Vainer2024CollaborativeCF} using diffusion models. Our work generates a textured object with homogeneous material properties from a single image under natural illumination at fast generation speeds.

\section{Method}

We propose SF3D, which converts a single object image into a textured and UV-unwrapped 3D model with de-lit albedo and material properties. As explained in the introduction section, with SF3D, we aim to fix the issues shown in \fig{issues} and introduce additional quality enhancements.

\inlinesection{Preliminaries.}
Our method is based on TripoSR~\cite{tochilkin2024triposr,hong2024lrm}, which trains a large transformer-based network that outputs a Triplane~\cite{Chan2021eg3d} based 3D representation from a single image. 
TripoSR is trained with multi-view image datasets without explicit 3D supervision. 
In TripoSR, the image is encoded using DINO~\cite{caron2021emerging} and passed through a Transformer network to generate a 3D triplane at a resolution of $64\times64$. The triplane features are then decoded into RGB colors and rendered using standard NeRF~\cite{mildenhall2020} rendering into multiple views for training. TripoSR only learns the view-independent colors and cannot model reflective objects. Several issues of the TripoSR (and other similar networks) are explained in the introduction (\fig{issues}).

\input{fig/overview}

\inlinesection{SF3D Overview}.
With SF3D, we propose several improvements to TripoSR~\cite{tochilkin2024triposr} to improve the output quality in different aspects. As illustrated in~\fig{overview}, SF3D has 5 main components: 
1. An enhanced transformer network that predicts higher resolution triplanes, which helps in reducing aliasing artifacts (top left in the figure);
2. A material estimation network (bottom left) predicts material properties, which helps handle an object's reflective properties.
3. Illumination prediction (bottom right) to tackle illumination disentangling, which helps in outputting homogeneous objects without shadows; 
4. Mesh extraction and refinement with the prediction of vertex offsets and surface normals (top right), which helps in smoother output shapes with fewer mesh extraction artifacts; and
5. A fast UV-unwrapping and export module (right) that helps produce low-poly meshes and high-resolution textures. Next, we explain each of these modules in detail.

\tightsubsection{Enhanced Transformer}
First, as illustrated in~\fig{overview}, we transitioned from DINO~\cite{caron2021emerging} used in TripoSR to the improved DINOv2~\cite{oquab2023dinov2} network to obtain image tokens for the transformer.
We observed that low-resolution ($64\times64$) triplanes used in TripoSR and other networks~\cite{hong2024lrm,openlrm} introduce noticeable artifacts, especially in scenarios with high-frequency and high-contrast texture patterns. \fig{aliasing_issue} (Middle) illustrates these aliasing artifacts. 
The triplane resolution is directly correlated with the presence of these artifacts, which we identify as an aliasing issue that can be mitigated by increasing the resolution. The increased capacity also improves the geometry.

\input{fig/aliasing_issue}

Naively increasing the triplane resolution quadratically increases the transformer complexity. We take inspiration from the recent PointInfinity~\cite{huang2024pointinfinity} work and propose an enhanced Transformer network that outputs higher-resolution triplanes. PointInfinity proposes an architecture where the complexity remains linear concerning the input size by avoiding the self-attention on the higher resolution triplane tokens. 
With this addition, we produce $96\times 96$ resolution triplanes with 1024 channels. We further increased the triplane resolution by shuffling the output features across dimensions, resulting in 40-channel features at $384 \times 384$ resolution. Further details about the architecture are available in the supplements. \fig{aliasing_issue} (Right) shows fewer aliasing artifacts with our higher resolution triplanes.

\tightsubsection{Material Estimation}
To enhance the output mesh appearance for the reflective objects, SF3D also outputs the material properties of metallic and roughness parameters.
Ideally, one would like to estimate the spatially varying material properties at 3D output locations, but this is an inherently challenging and ill-posed learning problem that requires many high-quality 3D data with spatially varying materials.
To overcome these challenges, we propose simplifying the material estimation problem by estimating a single metallic and roughness value for the entire object. Although this non-spatially varying material mainly applies to homogeneous objects, we find that it significantly improves the visual quality of our mesh predictions.
Specifically, as illustrated in~\fig{overview}, we propose `Material Net' that predicts the metallic and roughness values from the input image. 

For pre-training the Material Net, we selected a subset of 3D objects with PBR material properties from the synthetic training dataset and rendered them under different illuminations and viewpoints.
We observe that directly regressing the material values often leads to training collapse, where the network always predicts a roughness value of $0.5$ and a metallic value of $0$.
As a remedy, we propose a probabilistic prediction approach, where we predict the parameters of a Beta distribution and minimize the log-likelihood during training.
This stabilizes the training by allowing for uncertainty in this ambiguous material estimation task and prevents the collapse observed with direct regression.
During inference and training of \ours, we do not sample the distribution but calculate the mode of the distribution.

We implement the Material Net by first passing the image through the frozen CLIP image encoder~\cite{radford2021clip} to extract semantically meaningful latents and pass them through 2 separate MLPs with 3 hidden layers and 512 width to output the parameters for the distributions.

\tightsubsection{Illumination Modeling}
We propose explicitly estimating the illumination in the input image to account for varying shadings (e.g., shadows).
Otherwise, the 3D outputs would have baked-in illumination effects into their RGB colors, as illustrated in~\fig{issues}.
To this end, we propose a \emph{Light Net} (\fig{overview} bottom right) that predicts the spherical Gaussian (SG) illumination map from the estimated triplanes. The rationale here is that the triplanes encode the global structure and appearance of the input object and should account for the 3D spatial relationships and changes in illumination over the object surface. 
We use the $96\times96$ resolution triplanes from the transformer and pass them through 2 CNN layers, followed by a max pool and final MLP with three hidden layers and a feature dimension 512 for all layers.
Light Net outputs the grayscale amplitude values for 24 SGs with a Softplus activation to ensure positive values.
The axis and sharpness values for these SGs remain fixed and are set up to cover the entire sphere.
These amplitude values allow us to implement a deferred physically based rendering approach similar to that used in NeRD~\cite{Boss2021}. 

Our method also incorporates a lighting demodulation loss $\mathcal{L}_\text{Demod}$ during the training phase, inspired by the works of Hasselgren \etal~\cite{hasselgrenShapeLightMaterial2022} and Voleti \etal~\cite{voleti2024sv3d}. 
This loss function ensures that the lighting on an object with an entirely white albedo closely matches the luminance of the input image. 
The demodulation loss enforces consistency between the learned illumination and the lighting conditions observed in the training data. 
This can be seen as a bias to resolve the ambiguity between appearance and shading~\cite{Adelson1996}.

\tightsubsection{Mesh Extraction and Refinement}
\fig{overview} (top right) illustrates this module.
We convert the estimated triplanes into a mesh using a differentiable Marching Tetrahedron (DMTet)~\cite{shen2021dmtet} technique. As explained in the introduction (\fig{issues}), Marching Cubes (MC) usually results in several staircase artifacts on the resulting meshes.
We propose two new MLP heads to refine the meshes as a remedy. One predicts vertex offsets $\vect{v}_o \in \mathbb{R}^3$, and another one predicts the world space vertex normals $\vect{\hat{n}} \in \mathbb{R}^3$.
Inspired by MeshLRM~\cite{Wei2024MeshLRMLR}, we also implemented small split decoder MLPs for these two networks, which proved beneficial for performance and efficiency.
We found that the vertex offsets can reduce artifacts from the tetrahedral grids, and the world space normals can add details to the flat mesh triangles.
Given that the normal predictions are initially unreliable, we stabilize the training by using spherical linear interpolation (slerp) between the geometry normals $\vect{n} \in \mathbb{R}^3$ and our predictions. 
This slerp is used during the initial 5K training steps.

To regularize the mesh estimation, we use several training losses: a normal consistency loss $\mathcal{L}_\text{Nrm consistency}$, a Laplacian smoothness loss $\mathcal{L}_\text{Laplacian}$ as implemented by threestudio~\cite{threestudio2023}, and a vertex offset regularization $\mathcal{L}_\text{Offset}=\vect{v}_o^2$. For supervising the normal prediction, we use a geometry normal replication loss $\mathcal{L}_\text{Nrm repl}=1 - \vect{n} \cdot \vect{\hat{n}}$, where $\cdot$ is the dot product and a normal smoothness loss to ensure the smoothness of normal predictions in 3D. This loss is achieved by adding a small offset $\vect{\epsilon} \in \mathbb{R}^3$ around a query location $\vect{x} \in \mathbb{R}^3$. The loss is then defined as $\mathcal{L}_\text{Nrm smooth}=(\vect{\hat{n}}(\vect{x}) - \vect{\hat{n}}(\vect{x}+\vect{\epsilon}))^2$.

\input{fig/export_overview}

\tightsubsection{Fast UV-Unwrapping and Export}
The final stage of SF3D is an export pipeline that outputs the final 3D mesh along with the corresponding UV atlas. Our export pipeline follows multiple stages to ensure efficient and effective handling of 3D models. An overview of these stages is provided in \fig{export_overview}, where we first do fast UV-unwrapping. We then bake the world positions and occupancy to the UV atlas, which we use for querying the albedo and normal. This results in the final textured 3D mesh.
The entire export process only takes \SI{150}{\milli\second}.

\input{fig/uv_unwrapping_explained}

UV unwrapping is traditionally a computationally intensive process. Existing methods require several seconds for UV unwrapping, which is impractical when we are aiming for sub-second generation speeds.
To address this inefficiency, we propose a Cube projection-based unwrapping method. The key advantage of this approach is it is parallelizable: each face of the mesh can independently decide which cube face to project onto, based on its surface normal.

Our UV unwrapping process is illustrated in~\fig{uv_unwrapping}.
We initially align the output mesh based on the most dominant axes with the cube projection coordinate system. After each mesh face selects the appropriate cube direction, we address potential occlusions. Without managing occlusions, different faces could share the same UV coordinates, leading to artifacts in the texture. We detect occlusions in the UV atlas by performing 2D triangle-triangle intersection tests. We filter triangles by their proximity to the triangle centers to make the process efficient. If an intersection is detected, we sort the intersecting triangles based on their depth in the plane, keeping the first intersection and marking the others for reassignment to different UV atlas areas. The first intersection is placed in the top third of the UV atlas and the second intersection is placed in the bottom left area.
The remaining triangles are organized into a grid in the bottom right section of the atlas.
We also rotate each island to minimize the shading seams by following radial z tangent orientation.
We then assign each face to a position in the UV atlas, as illustrated in \fig{uv_unwrapping}.

Next, we bake the world positions and the occupancy data with UV-unwrapping into the final UV atlas. This allows us to query the world positions within the chart from our triplanes and decode the albedo and surface normals into additional textures. We transform the world-space normal map into a tangent-space normal map using the tangent and bitangent vectors.
We add margins to the UV atlases to prevent visible seams at UV island borders. This is achieved through an iterative process: in each iteration, we perform a $3\times3$ partial convolution based on the occupied areas, using the valid regions of the kernel. We then use a $3\times3$ max pooling operation to expand the occupied regions of the UV atlas, placing the mean values in the newly expanded areas while preserving the original regions. This iterative extension ensures that the textures smoothly blend outwards.

We incorporate our image estimator's metallic and roughness values and pack everything into a GLB file, ready for efficient rendering and use in various applications.

\tightsubsection{Overall Training and Loss Functions}
Directly training our method with mesh rendering yielded unsatisfactory results. Hence, we pre-trained it on the NeRF task. Following this pre-training, we transitioned to mesh training, replacing the NeRF rendering with differentiable mesh rendering and SG-based shading. Given the introduction of light estimation, we found that using larger batch sizes aids convergence.
We initiate training with a batch size of 192 and a rendering resolution of $128\times128$, training for 10K steps. In the subsequent stage, we reduce the batch size to 128 and increase the resolution to $256\times256$, continuing for 20K steps. The final stage involves 80K steps at a $512\times512$ resolution with a batch size of 96.

The loss functions remain consistent across all mesh training stages. 
We primarily use image-based metrics to compare our rendered and shaded reconstructions $\vect{\hat{I}}$ with the GT image $\vect{I}$. 
These include MSE $\mathcal{L}_\text{MSE}$ and LPIPS~\cite{zhang2018perceptual} $\mathcal{L}_\text{LPIPS}$ losses. 
We also incorporate a mask loss $\mathcal{L}_\text{Mask}$ between the GT mask $M$ and the predicted opacity $\hat{M}$, defined as an MSE loss. We then define three loss formulations for the rendering, mesh regularization, and shading:

\begin{tiny}
\begin{align}
    \mathcal{L}_\text{render} &= 
    \underbrace{\lambda_\text{MSE}}_{10} \mathcal{L}_\text{MSE} + 
    \underbrace{\lambda_\text{LPIPS}}_{2} \mathcal{L}_\text{LPIPS} +
    \underbrace{\lambda_\text{Mask}}_{10}\mathcal{L}_\text{Mask} \\
    \mathcal{L}_\text{mesh} &= \underbrace{\lambda_\text{Laplacian}}_{0.01}\mathcal{L}_\text{Laplacian} + 
    \underbrace{\lambda_\text{Nrm Consistency}}_{0.001}\mathcal{L}_\text{Nrm consistency} + 
    \underbrace{\lambda_\text{Offset}}_{0.1} \mathcal{L}_\text{Offset} \\
    \mathcal{L}_\text{shading} &= 
    \underbrace{\lambda_\text{Nrm repl}}_{0.2} \mathcal{L}_\text{Nrm repl} 
    \underbrace{\lambda_\text{Nrm smooth}}_{0.02}\mathcal{L}_\text{Nrm smooth} + \underbrace{\lambda_\text{Demod}}_{0.01}\mathcal{L}_\text{Demod}
\end{align}
\end{tiny}

\noindent The total loss is defined as:

\begin{tiny}
\begin{equation}
    \mathcal{L} = \mathcal{L}_\text{render} + \mathcal{L}_\text{mesh} + \mathcal{L}_\text{shading}
\end{equation}
\end{tiny}
\vspace{-3mm}
\section{Results}

\begin{figure*}[ht!]
    \centering
    \resizebox{0.85\linewidth}{!}{ 
        \renewcommand{\arraystretch}{0.6}%
        \setlength{\tabcolsep}{0pt}%
        \begin{tabular}{%
            @{}%
            >{\centering\arraybackslash}m{2cm}@{\hskip 0.25cm}|@{\hskip 0.25cm}%
            >{\centering\arraybackslash}m{2cm}%
            >{\centering\arraybackslash}m{2cm}@{\hskip 0.25cm}|@{\hskip 0.25cm}%
            >{\centering\arraybackslash}m{2cm}%
            >{\centering\arraybackslash}m{2cm}@{\hskip 0.5cm}%
            >{\centering\arraybackslash}m{2cm}%
            >{\centering\arraybackslash}m{2cm}@{\hskip 0.5cm}%
            >{\centering\arraybackslash}m{2cm}%
            >{\centering\arraybackslash}m{2cm}@{\hskip 0.5cm}%
            >{\centering\arraybackslash}m{2cm}%
            >{\centering\arraybackslash}m{2cm}@{\hskip 0.5cm}%
            >{\centering\arraybackslash}m{2cm}%
            >{\centering\arraybackslash}m{2cm}@{\hskip 0.5cm}%
            >{\centering\arraybackslash}m{2cm}%
            >{\centering\arraybackslash}m{2cm}
            @{}%
        }
    Input & \multicolumn{2}{c}{GT} & \multicolumn{2}{c}{CRM} & \multicolumn{2}{c}{LGM} & \multicolumn{2}{c}{InstantMesh} & 
    \multicolumn{2}{c}{TripoSR} & \multicolumn{2}{c}{Ours} \\
\includegraphics[width=2cm]{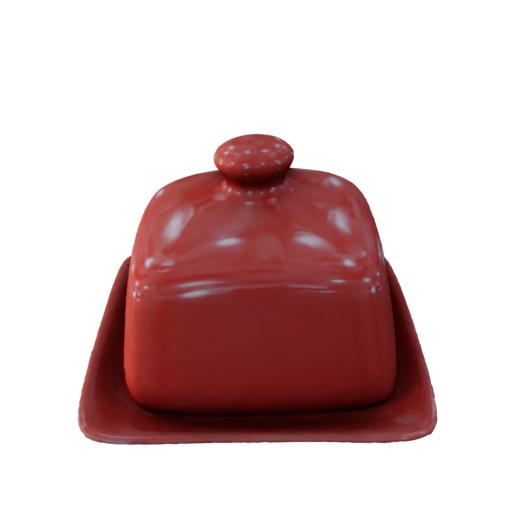} & \includegraphics[width=2cm]{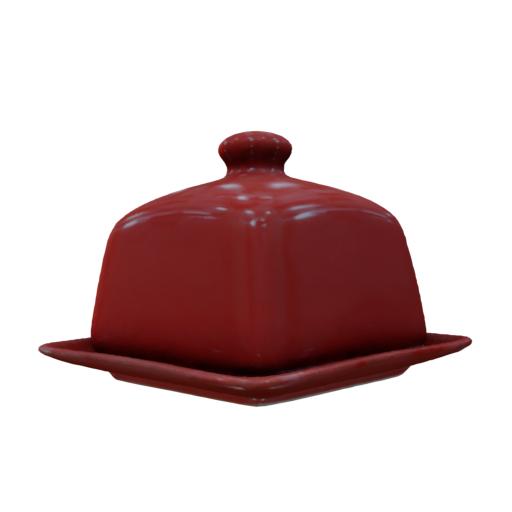} & \includegraphics[width=2cm]{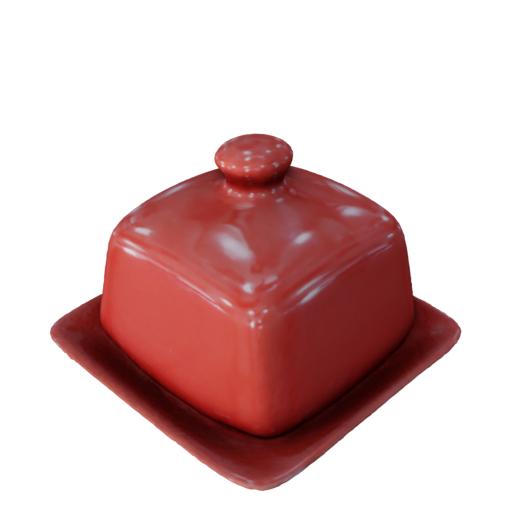} & \includegraphics[width=2cm]{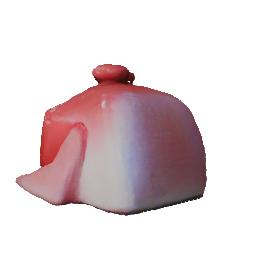} & \includegraphics[width=2cm]{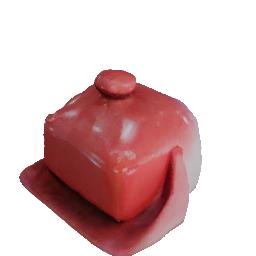} & \includegraphics[width=2cm]{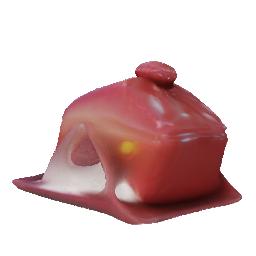} & \includegraphics[width=2cm]{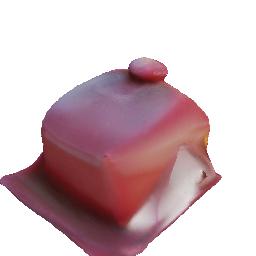} & \includegraphics[width=2cm]{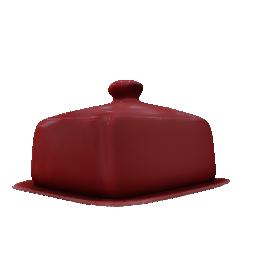} & \includegraphics[width=2cm]{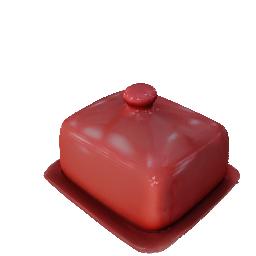} & \includegraphics[width=2cm]{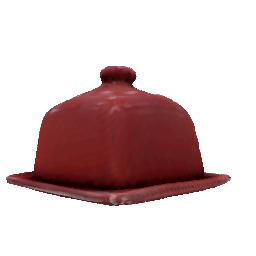} & \includegraphics[width=2cm]{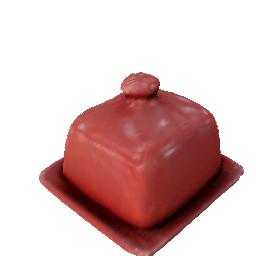} & \includegraphics[width=2cm]{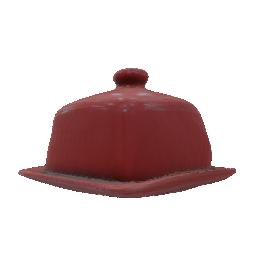} & \includegraphics[width=2cm]{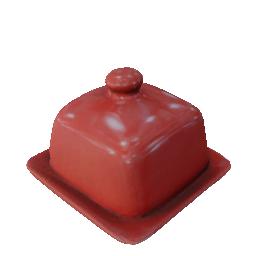} \\
\includegraphics[width=2cm]{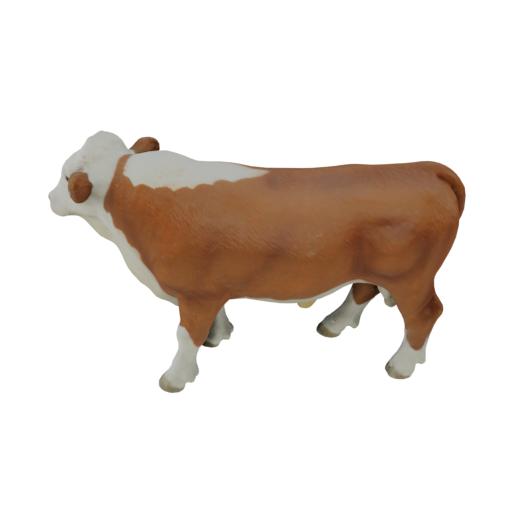} & \includegraphics[width=2cm]{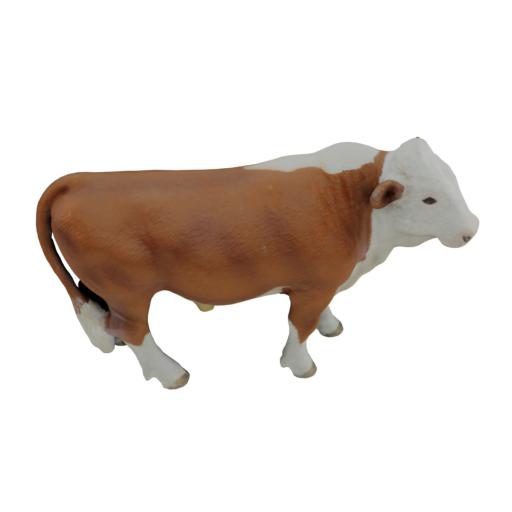} & \includegraphics[width=2cm]{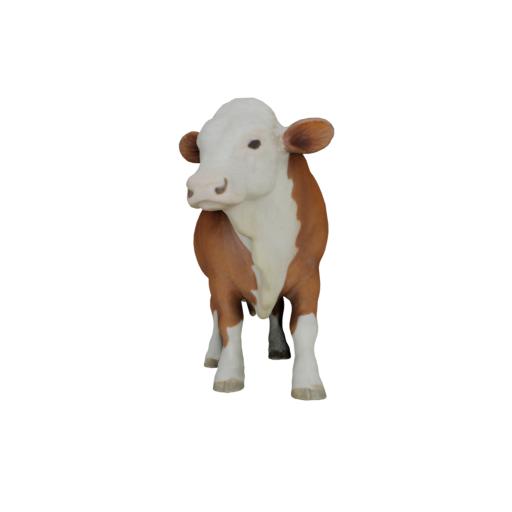} & \includegraphics[width=2cm]{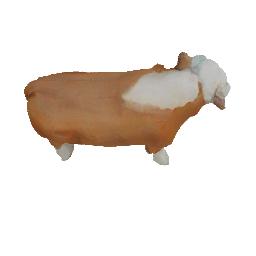} & \includegraphics[width=2cm]{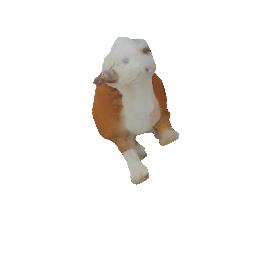} & \includegraphics[width=2cm]{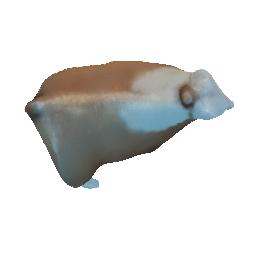} & \includegraphics[width=2cm]{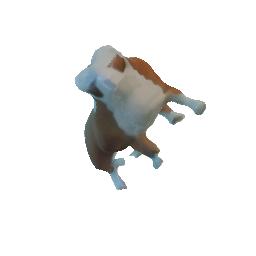} & \includegraphics[width=2cm]{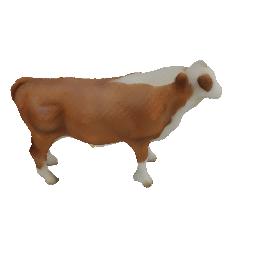} & \includegraphics[width=2cm]{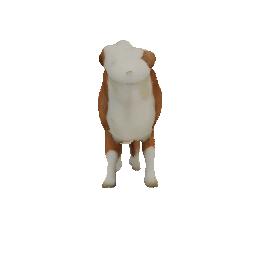} & \includegraphics[width=2cm]{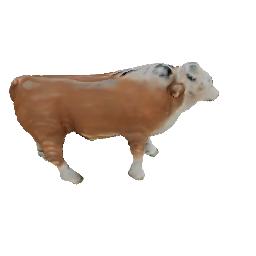} & \includegraphics[width=2cm]{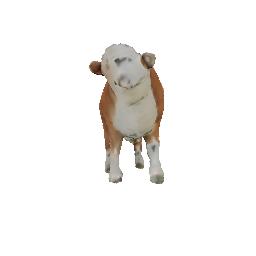} & \includegraphics[width=2cm]{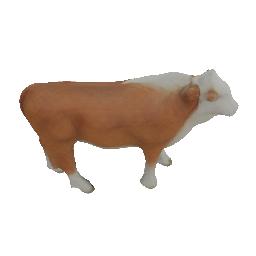} & \includegraphics[width=2cm]{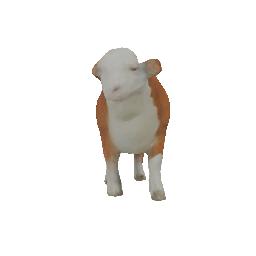} \\
\includegraphics[width=2cm]{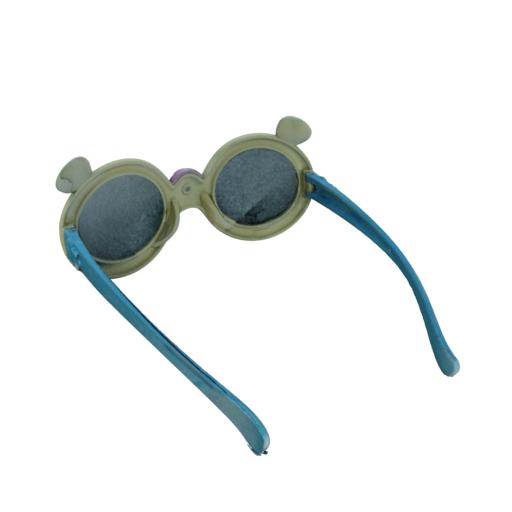} & \includegraphics[width=2cm]{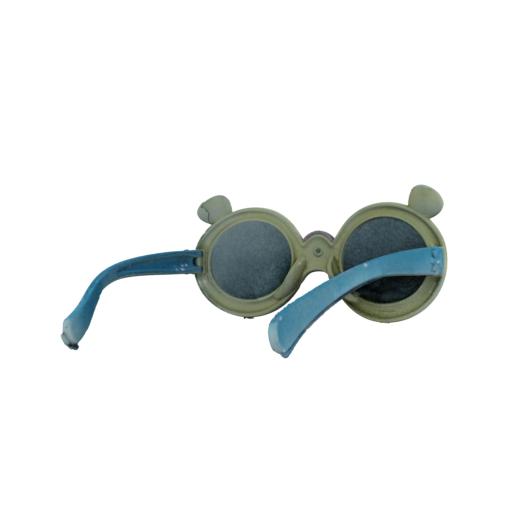} & \includegraphics[width=2cm]{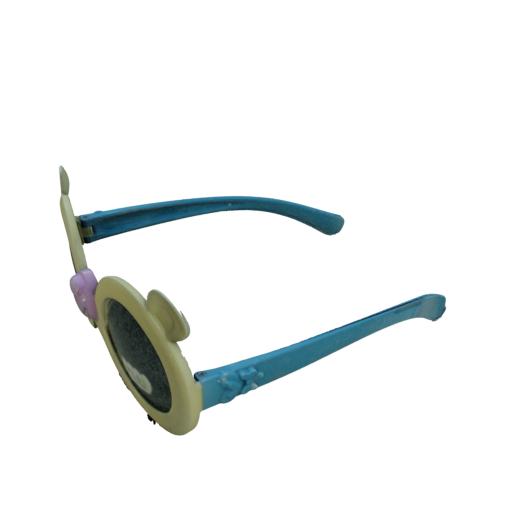} & \includegraphics[width=2cm]{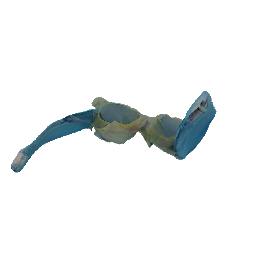} & \includegraphics[width=2cm]{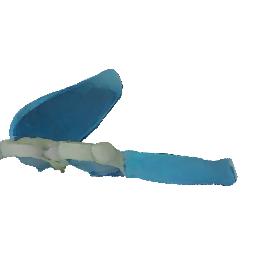} & \includegraphics[width=2cm]{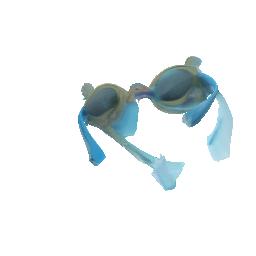} & \includegraphics[width=2cm]{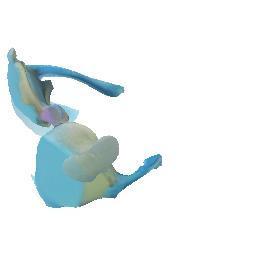} & \includegraphics[width=2cm]{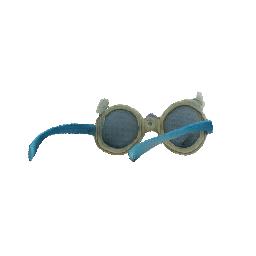} & \includegraphics[width=2cm]{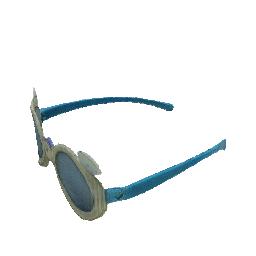} & \includegraphics[width=2cm]{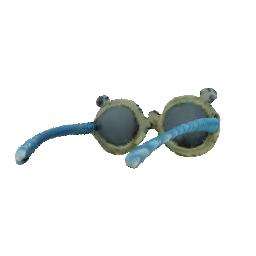} & \includegraphics[width=2cm]{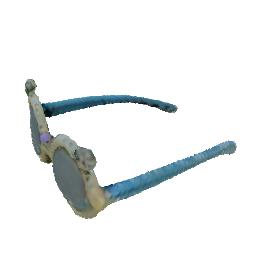} & \includegraphics[width=2cm]{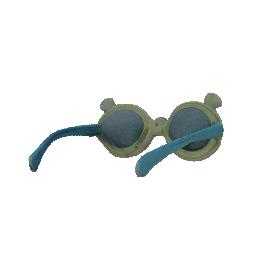} & \includegraphics[width=2cm]{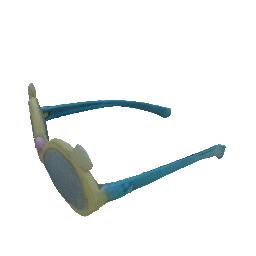} \\
\includegraphics[width=2cm]{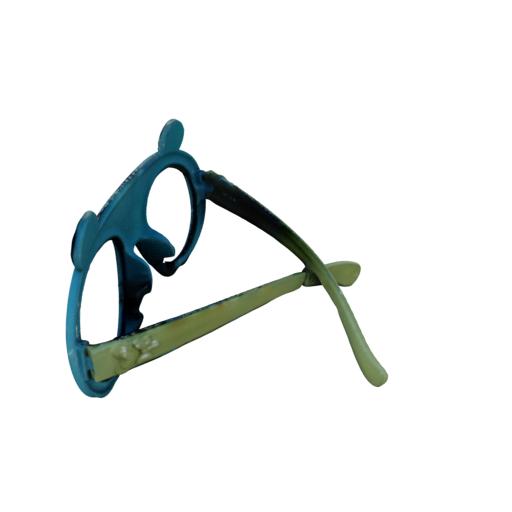} & \includegraphics[width=2cm]{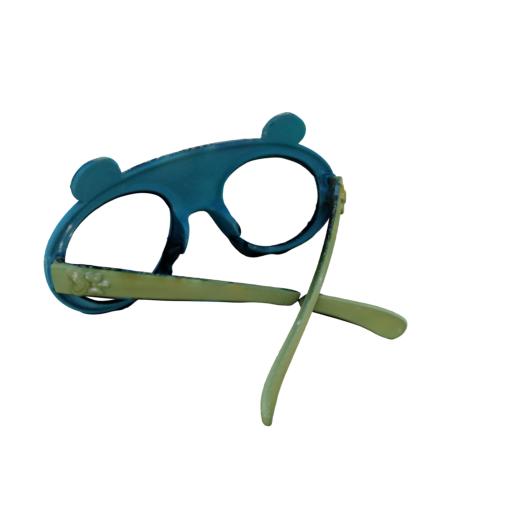} & \includegraphics[width=2cm]{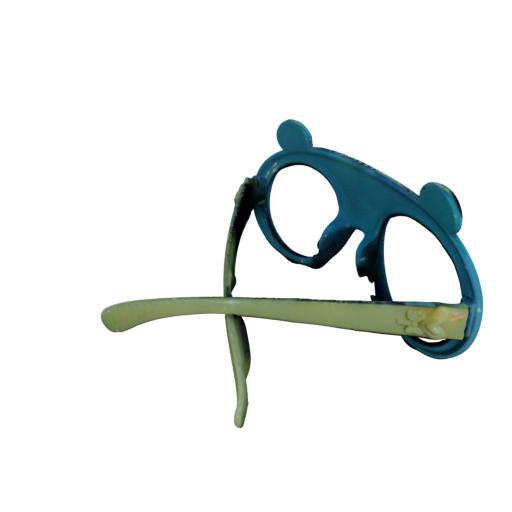} & \includegraphics[width=2cm]{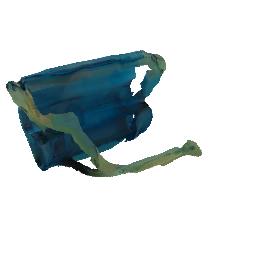} & \includegraphics[width=2cm]{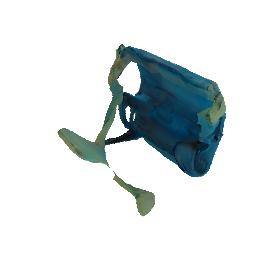} & \includegraphics[width=2cm]{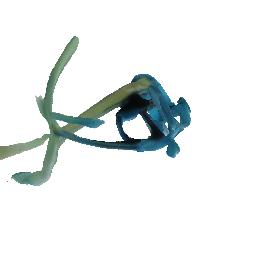} & \includegraphics[width=2cm]{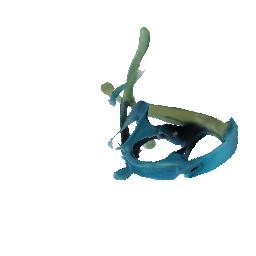} & \includegraphics[width=2cm]{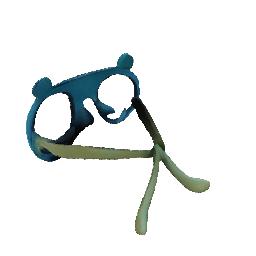} & \includegraphics[width=2cm]{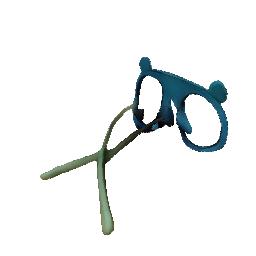} & \includegraphics[width=2cm]{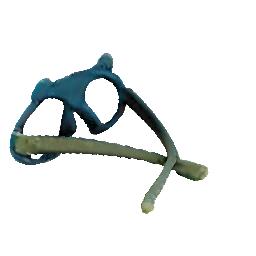} & \includegraphics[width=2cm]{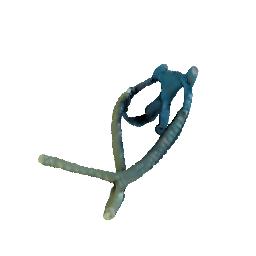} & \includegraphics[width=2cm]{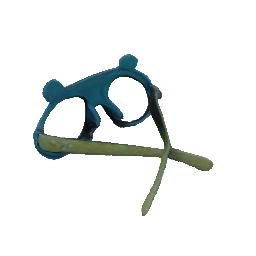} & \includegraphics[width=2cm]{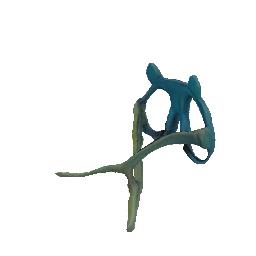} \\
\includegraphics[width=2cm]{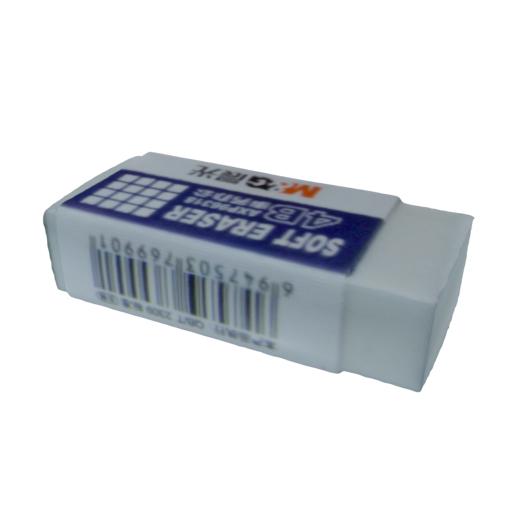} & \includegraphics[width=2cm]{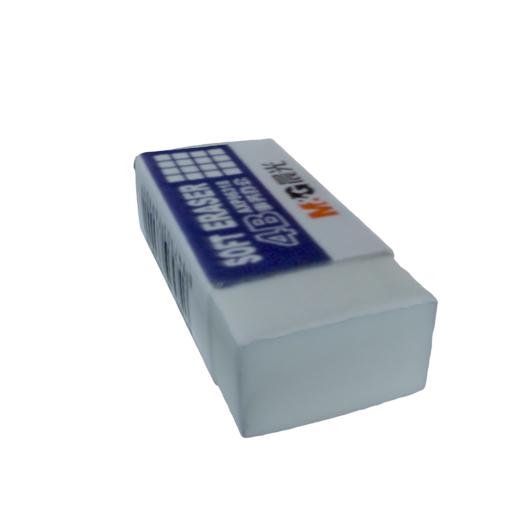} & \includegraphics[width=2cm]{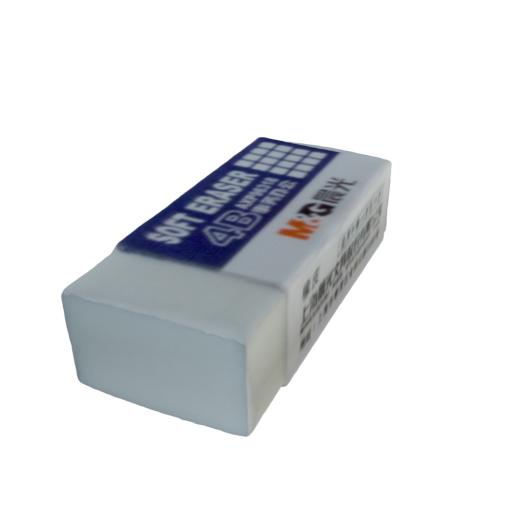} & \includegraphics[width=2cm]{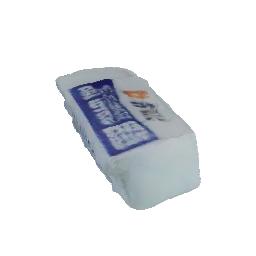} & \includegraphics[width=2cm]{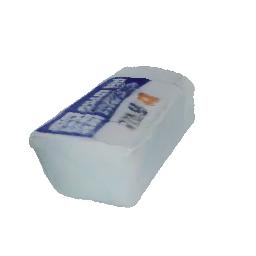} & \includegraphics[width=2cm]{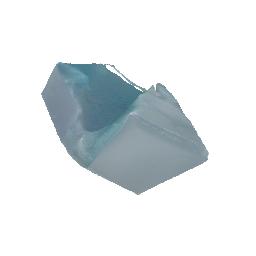} & \includegraphics[width=2cm]{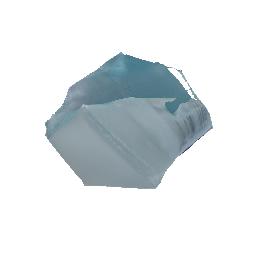} & \includegraphics[width=2cm]{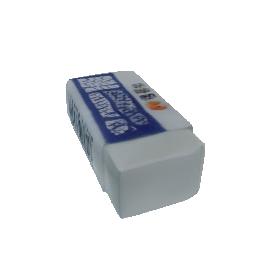} & \includegraphics[width=2cm]{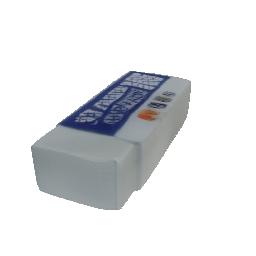} & \includegraphics[width=2cm]{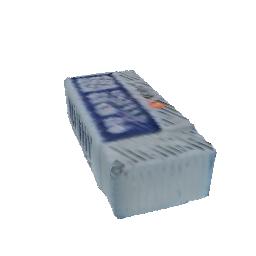} & \includegraphics[width=2cm]{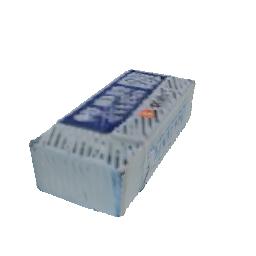} & \includegraphics[width=2cm]{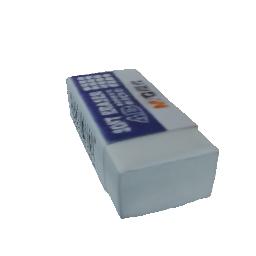} & \includegraphics[width=2cm]{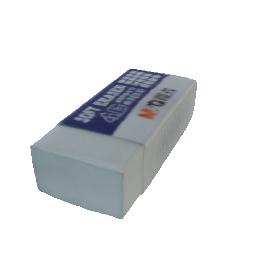} \\
\includegraphics[width=2cm]{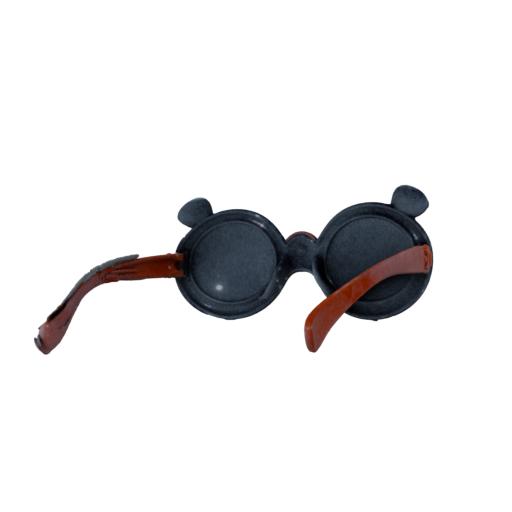} & \includegraphics[width=2cm]{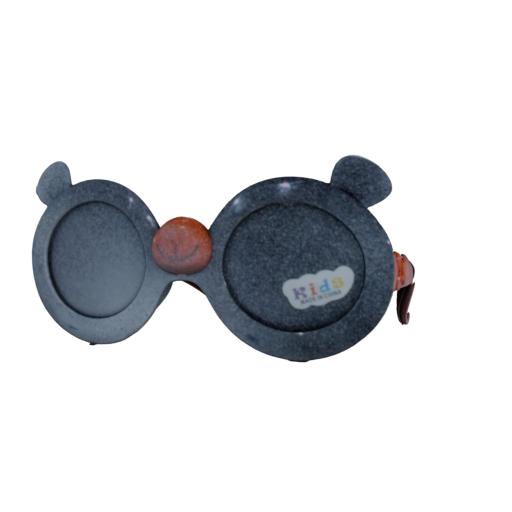} & \includegraphics[width=2cm]{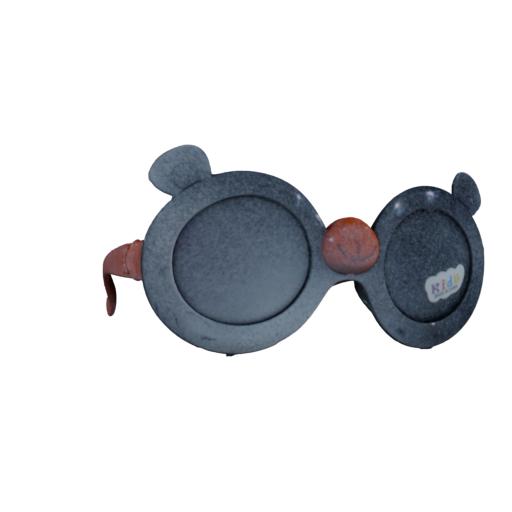} & \includegraphics[width=2cm]{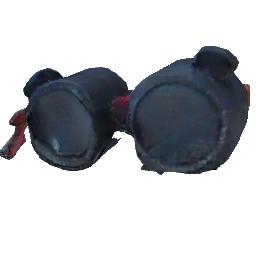} & \includegraphics[width=2cm]{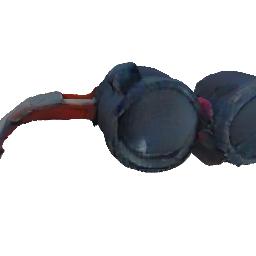} & \includegraphics[width=2cm]{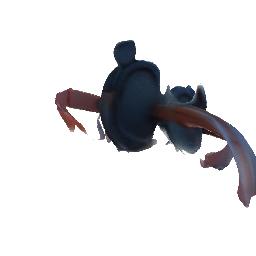} & \includegraphics[width=2cm]{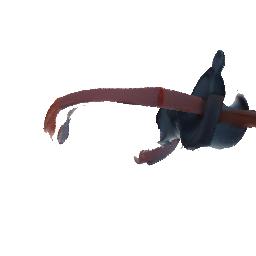} & \includegraphics[width=2cm]{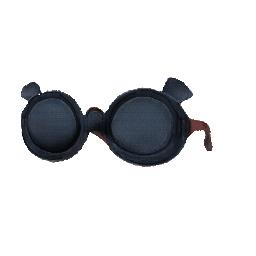} & \includegraphics[width=2cm]{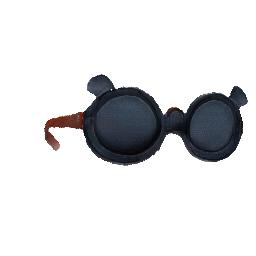} & 
\includegraphics[width=2cm]{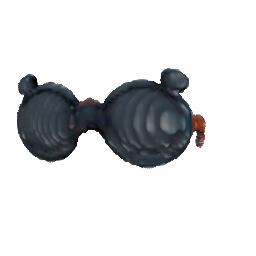} & \includegraphics[width=2cm]{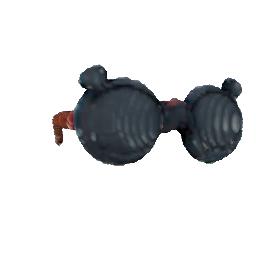} & \includegraphics[width=2cm]{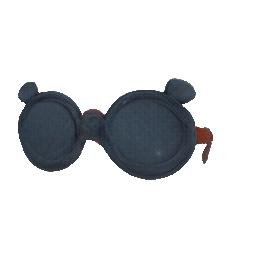} & \includegraphics[width=2cm]{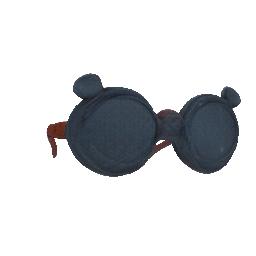} \\
\includegraphics[width=2cm]{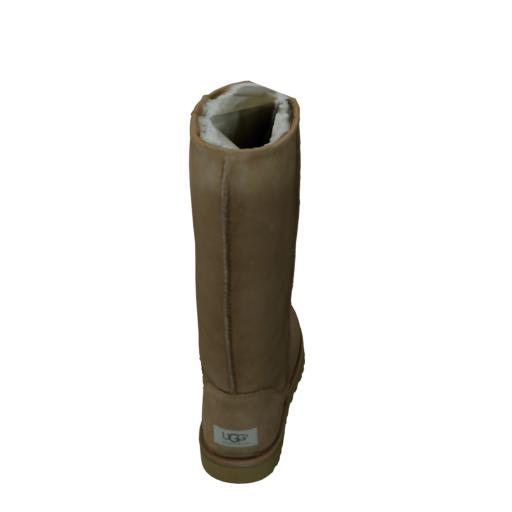} & \includegraphics[width=2cm]{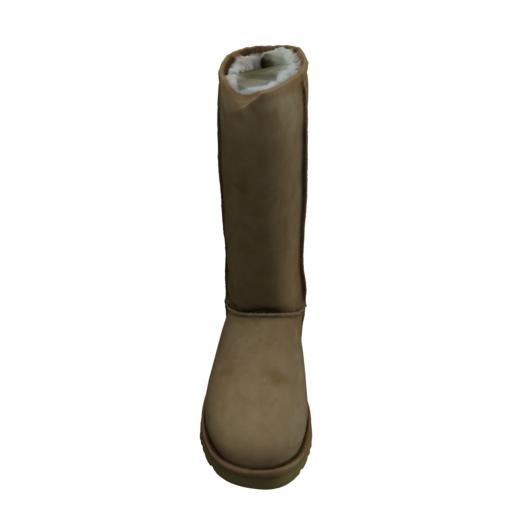} & \includegraphics[width=2cm]{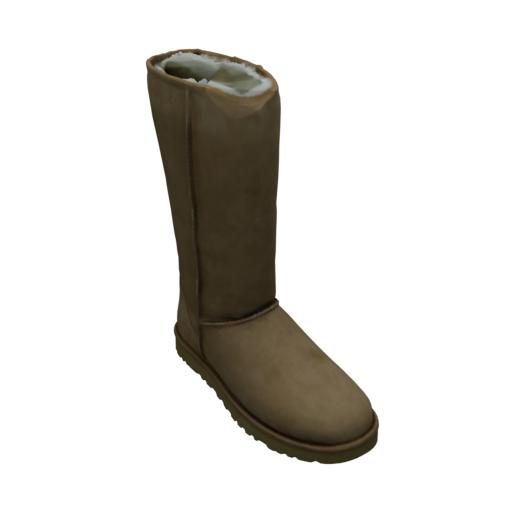} & \includegraphics[width=2cm]{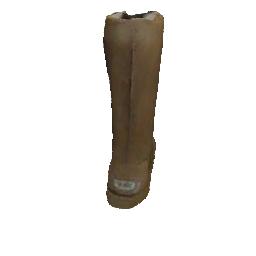} & \includegraphics[width=2cm]{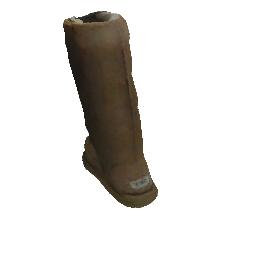} & \includegraphics[width=2cm]{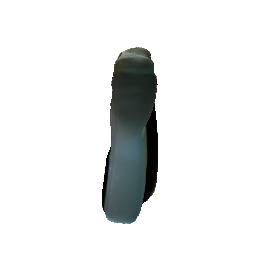} & \includegraphics[width=2cm]{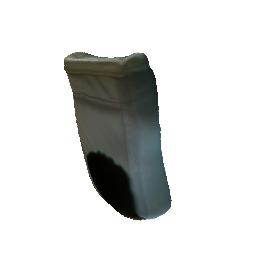} & \includegraphics[width=2cm]{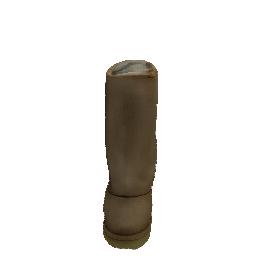} & \includegraphics[width=2cm]{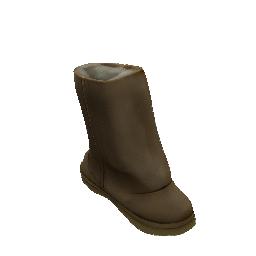} & \includegraphics[width=2cm]{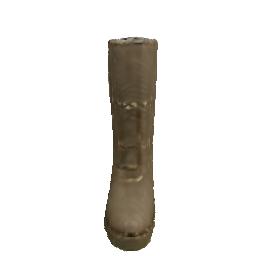} & \includegraphics[width=2cm]{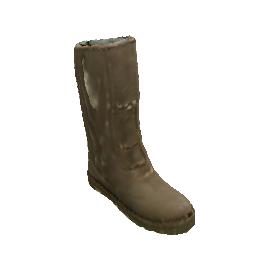} & \includegraphics[width=2cm]{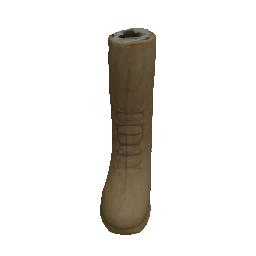} & \includegraphics[width=2cm]{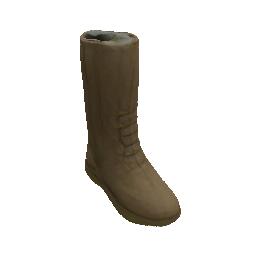} \\
\includegraphics[width=2cm]{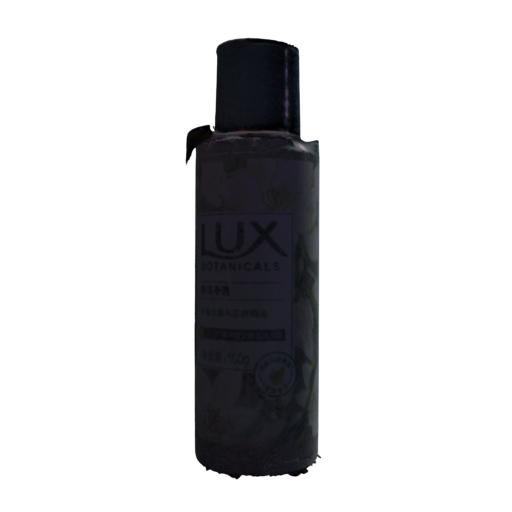} & \includegraphics[width=2cm]{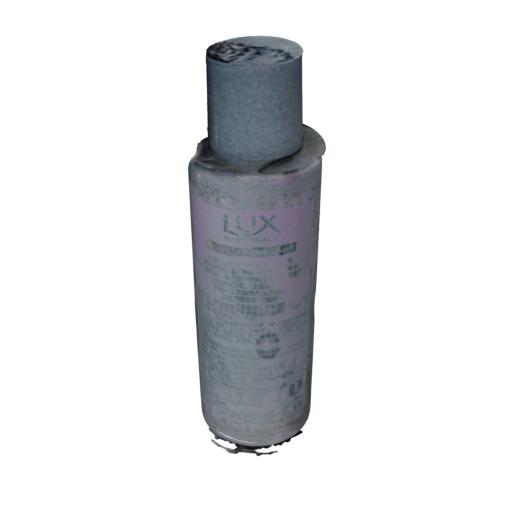} & \includegraphics[width=2cm]{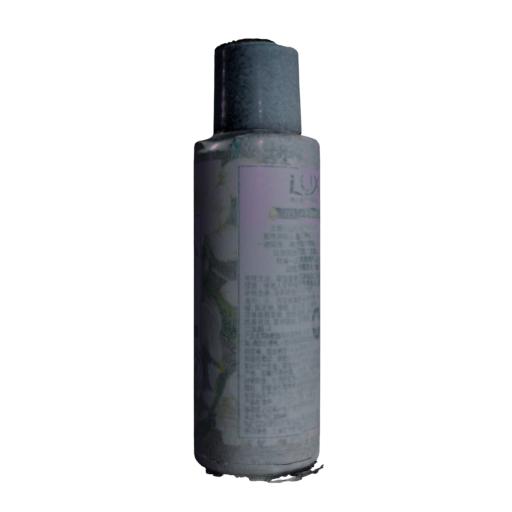} & \includegraphics[width=2cm]{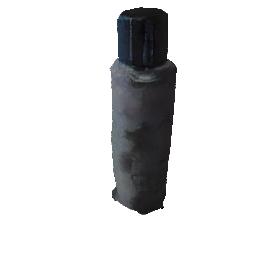} & \includegraphics[width=2cm]{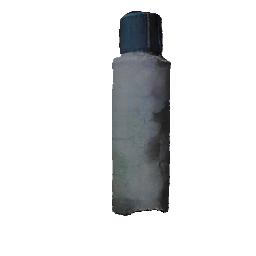} & \includegraphics[width=2cm]{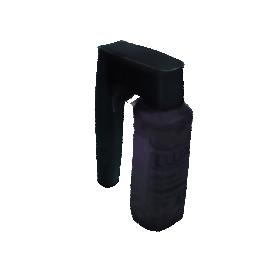} & \includegraphics[width=2cm]{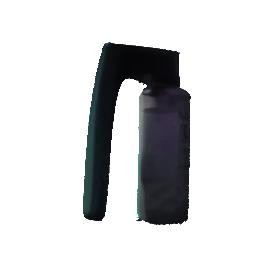} & \includegraphics[width=2cm]{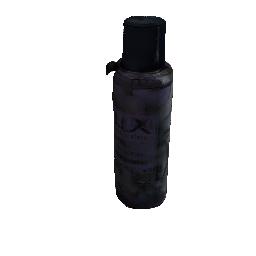} & \includegraphics[width=2cm]{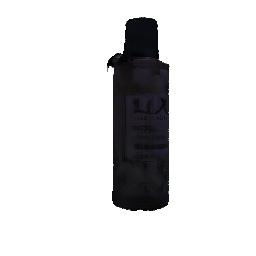} & \includegraphics[width=2cm]{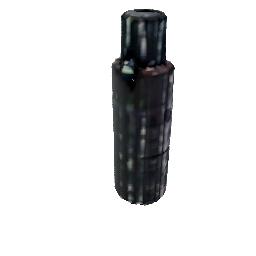} & \includegraphics[width=2cm]{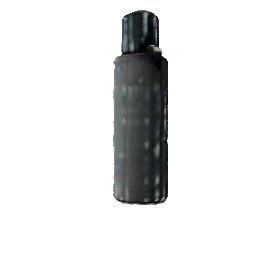} & \includegraphics[width=2cm]{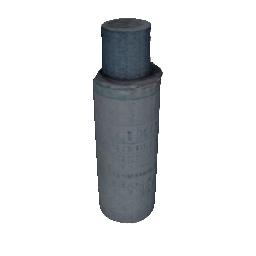} & \includegraphics[width=2cm]{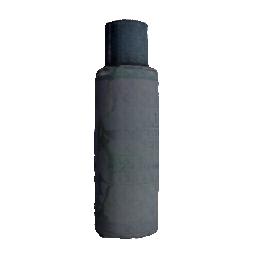} \\
\includegraphics[width=2cm]{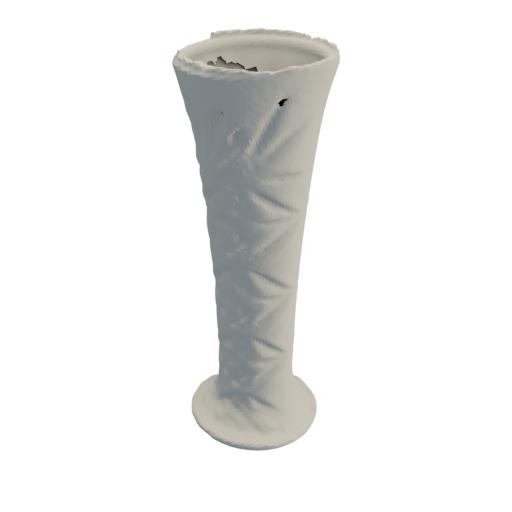} & \includegraphics[width=2cm]{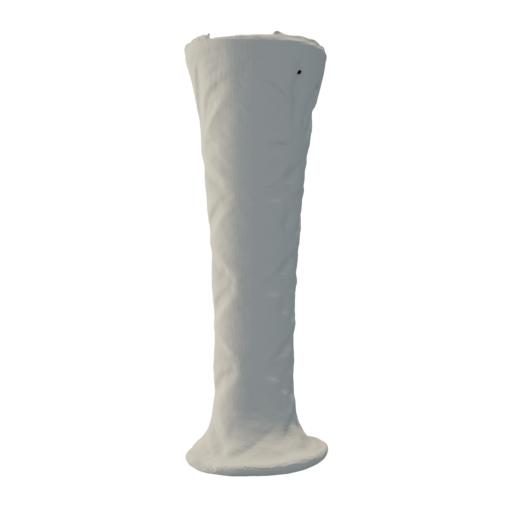} & \includegraphics[width=2cm]{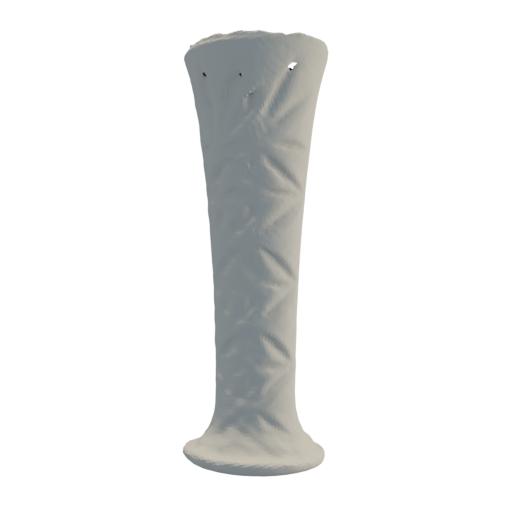} & \includegraphics[width=2cm]{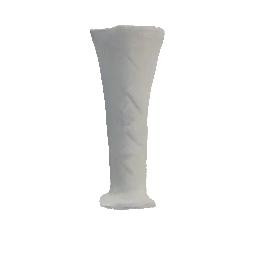} & \includegraphics[width=2cm]{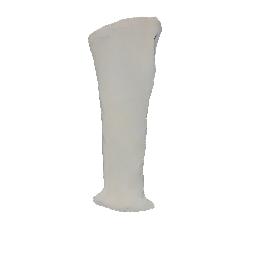} & \includegraphics[width=2cm]{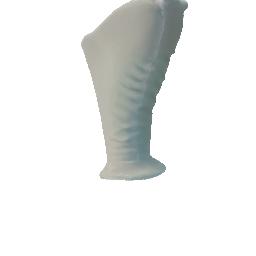} & \includegraphics[width=2cm]{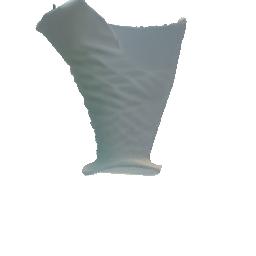} & \includegraphics[width=2cm]{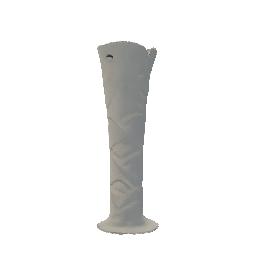} & \includegraphics[width=2cm]{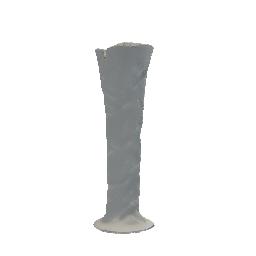} & \includegraphics[width=2cm]{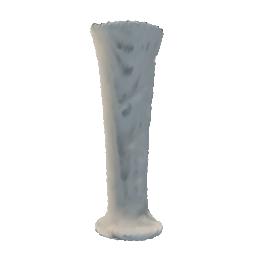} & \includegraphics[width=2cm]{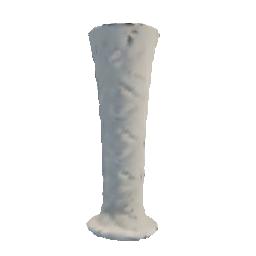} & \includegraphics[width=2cm]{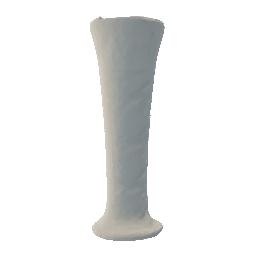} & \includegraphics[width=2cm]{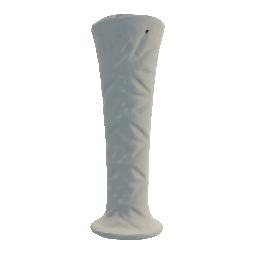} \\
\includegraphics[width=2cm]{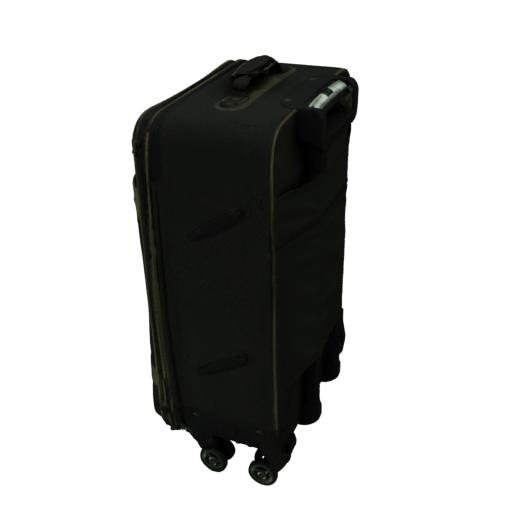} & \includegraphics[width=2cm]{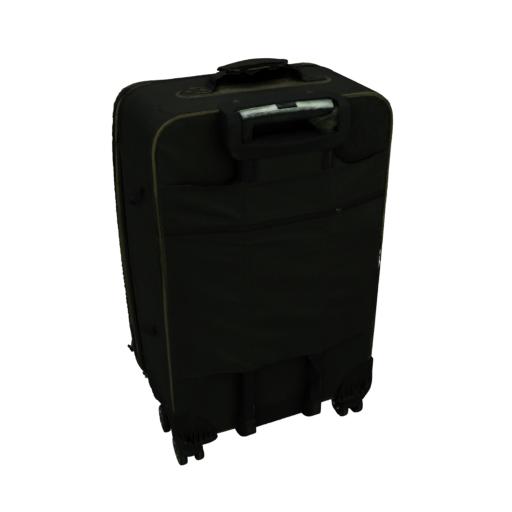} & \includegraphics[width=2cm]{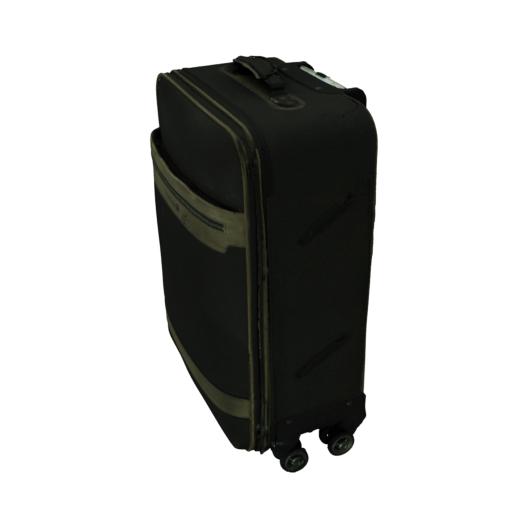} & \includegraphics[width=2cm]{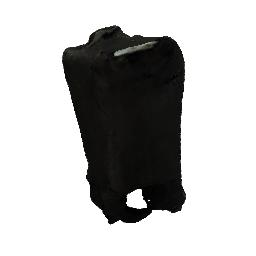} & \includegraphics[width=2cm]{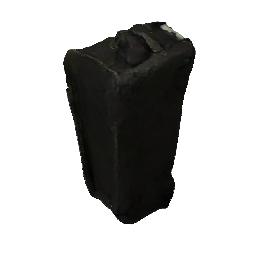} & \includegraphics[width=2cm]{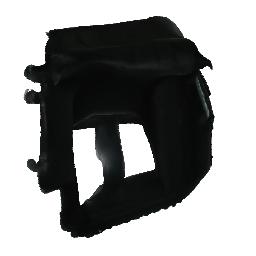} & \includegraphics[width=2cm]{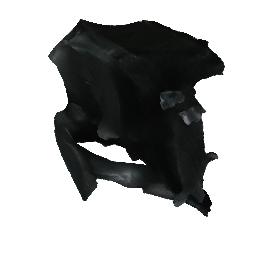} & \includegraphics[width=2cm]{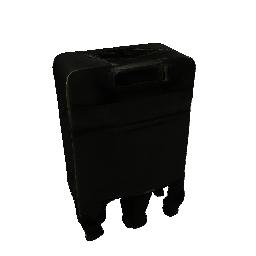} & \includegraphics[width=2cm]{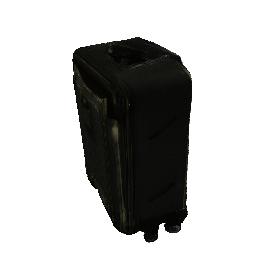} & \includegraphics[width=2cm]{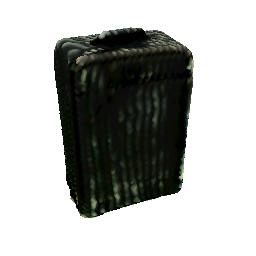} & \includegraphics[width=2cm]{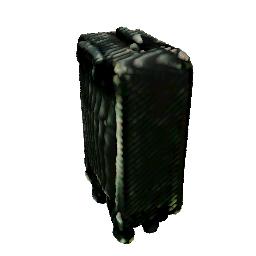} & \includegraphics[width=2cm]{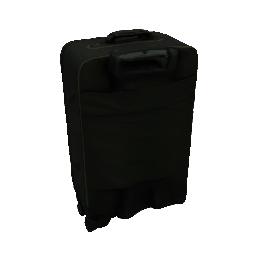} & \includegraphics[width=2cm]{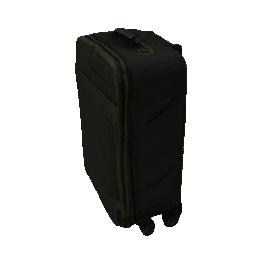} \\

        \end{tabular}
    }
    
\vspace{-2mm}
\titlecaptionof{figure}{Comparison on GSO and OmniObject3D}{Notice how our reconstructions produce consistent results with detailed textures and smooth shading.}
    \label{fig:multi_comparison_gso_omniobj}
\end{figure*}

\begin{table*}[ht!]
\centering
\resizebox{0.8\linewidth}{!}{ %
    \begin{tabular}{lccccccccc}
           & & \multicolumn{4}{c}{GSO} & \multicolumn{4}{c}{OmniObject} \\
           \cmidrule(lr){3-6} \cmidrule{7-10}
        Method & Time [s]\textdownarrow & CD\textdownarrow & FS$@$0.1\textuparrow & FS$@$0.2\textuparrow & FS$@$0.5\textuparrow & CD\textdownarrow & FS$@$0.1\textuparrow & FS$@$0.2\textuparrow & FS$@$0.5\textuparrow \\
        \midrule
        ZeroShape~\cite{Huang2023ZeroShapeRZ} & 0.9 & 0.160 & 0.489 & 0.759 & 0.952 & 0.144 & 0.507 & 0.786 & 0.969 \\
        OpenLRM~\cite{openlrm}               & 2.0 & 0.160 & 0.472 & 0.751 & 0.954 & 0.139 & 0.521 & 0.798 & 0.971 \\
        TripoSR~\cite{tochilkin2024triposr}  & \best{0.3} & \secondbest{0.111} & \secondbest{0.645} & \secondbest{0.869} & \secondbest{0.980} & \secondbest{0.103} & \secondbest{0.672} & \secondbest{0.889} & \secondbest{0.986}\\
        LGM~\cite{Tang2024LGMLM} & 64.6 & 0.195 & 0.376 & 0.654 & 0.928 & 0.205 & 0.344 & 0.631 & 0.921 
        \\
        CRM~\cite{Wang2024CRMSI} & 10.2 & 0.179 & 0.411 & 0.699 & 0.945 & 0.158 & 0.469 & 0.752 & 0.960 \\
        InstantMesh~\cite{Xu2024InstantMeshE3} & 32.4 & 0.138 & 0.549 & 0.801 & 0.967 & 0.138 & 0.560 & 0.811 & 0.964 \\

        \midrule
        SF3D (Ours)            & \secondbest{0.5} & \best{0.098} & \best{0.701} & \best{0.894} & \best{0.988} & \best{0.090} & \best{0.726} & \best{0.920} & \best{0.990} \\

    \end{tabular}
}
\vspace{-1mm}
\titlecaption{Comparison on 3D Metrics Demonstrating State-of-the-art Performance of SF3D}{It is worth pointing out that all other methods produce meshes with drastically higher polygon counts. This allows them to follow the surfaces more closely than our low polygon meshes. Our meshes can outperform the others by using our learned vertex offsets. Our method is also one of fastest to generate a mesh from an image due to our efficient extraction pipeline.
}
\label{tab:comparison_3d}
\end{table*}

\inlinesection{Datasets.}
For comparisons, we select GSO~\cite{downs2022gso} and OmniObject3D~\cite{wu2023omniobject3d} as our primary datasets for evaluation. 
We select 278 random scenes from GSO and 308 scenes from OmniObject3D for comparison. 
We render 16 views around the object and select a frontal as the conditioning view.

\inlinesection{Baselines.}
We compare with several recent methods for fast 3D  object reconstruction from a single image. 
We mainly focus on fast reconstruction models to maintain a consistent evaluation protocol across different techniques. 
We also mainly consider meshes as outputs and perform the evaluations on the mesh. 
We use the official implementation for all the baslines, and we evaluate all the methods under the same protocol.
We selected several recent and concurrent works with source code releases for comparisons. 
Specifically, we compare against OpenLRM~\cite{openlrm}, TripoSR~\cite{tochilkin2024triposr}, LGM~\cite{Tang2024LGMLM}, CRM~\cite{Wang2024CRMSI}, InstantMesh~\cite{Xu2024InstantMeshE3}, and ZeroShape~\cite{Huang2023ZeroShapeRZ}.
For OpenLRM, we selected the large Objaverse 1.1 model. 
We use H100 GPU for comparisons.

\begin{figure}[ht!]
    \centering
\resizebox{.9\linewidth}{!}{ 
\renewcommand{\arraystretch}{0.6}%
\setlength{\tabcolsep}{0pt}%
\begin{tabular}{%
    @{}%
    l
    >{\centering\arraybackslash}m{3cm}%
    >{\centering\arraybackslash}m{3cm}%
    >{\centering\arraybackslash}m{3cm}%
    >{\centering\arraybackslash}m{3cm}%
    >{\centering\arraybackslash}m{3cm}%
    >{\centering\arraybackslash}m{3cm}%
    @{}%
}
& Render & Diffuse & Roughness-Metallic & Normal & Relight 1 & Relight 2 \\[0cm]

\rotatebox[origin=c]{90}{GT} & 
\includegraphics[width=3cm]{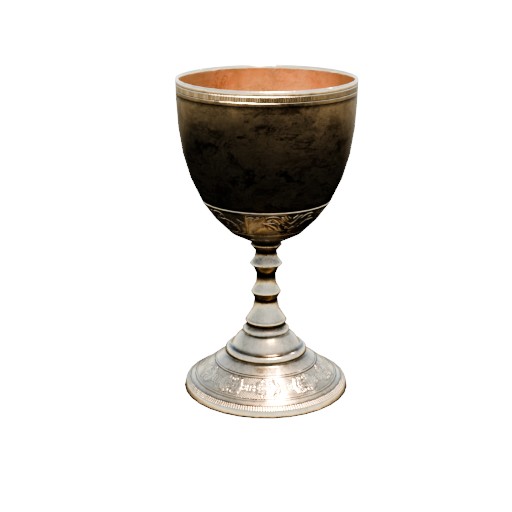} &
\includegraphics[width=3cm]{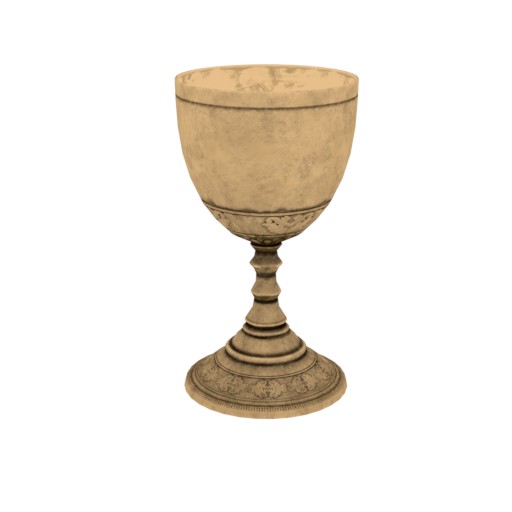} &
\includegraphics[width=3cm]{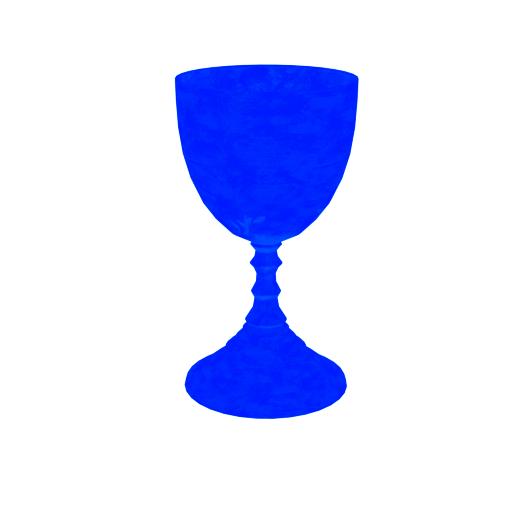} &
\includegraphics[width=3cm]{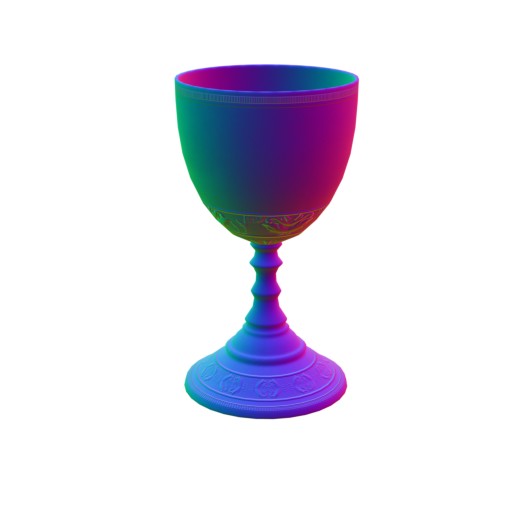} &
\includegraphics[width=3cm]{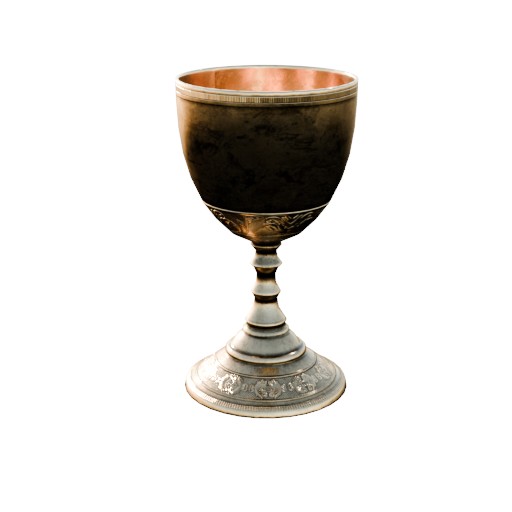} &
\includegraphics[width=3cm]{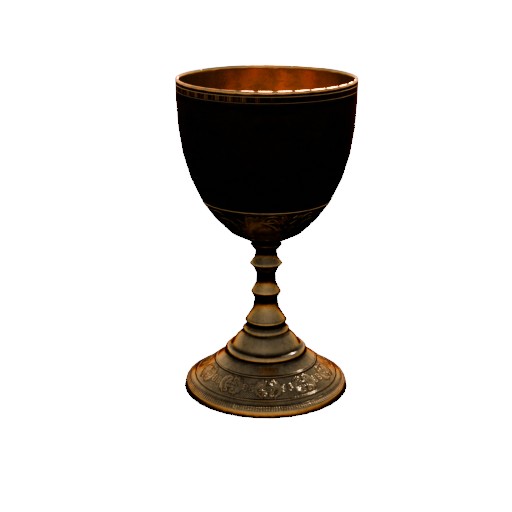} \\[-0.5cm]

\rotatebox[origin=c]{90}{Ours} & 
\includegraphics[width=3cm]{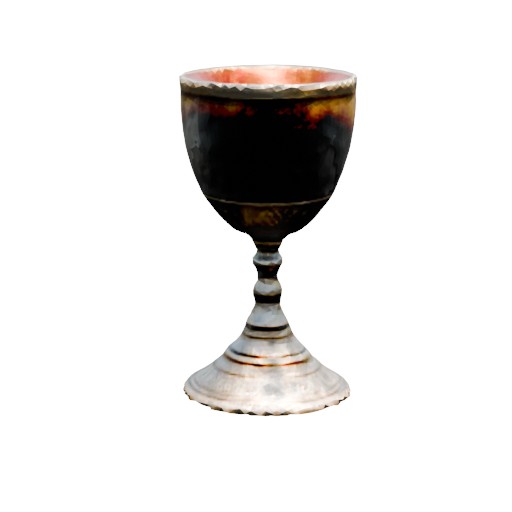} &
\includegraphics[width=3cm]{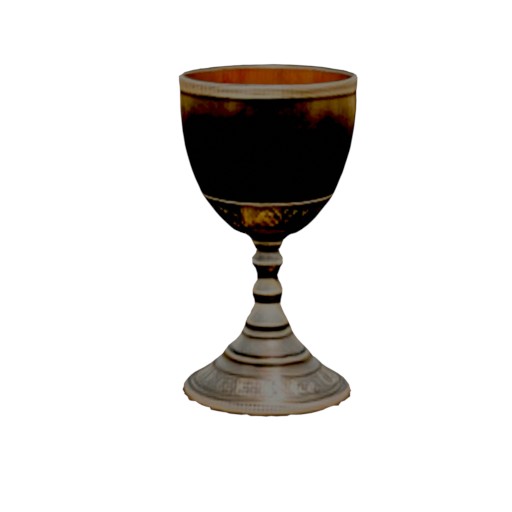} &
\includegraphics[width=3cm]{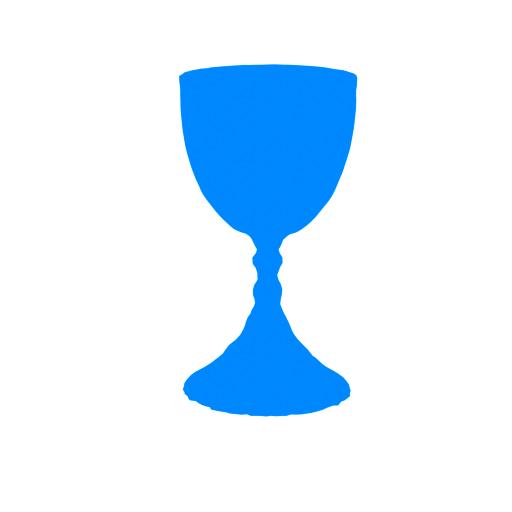} &
\includegraphics[width=3cm]{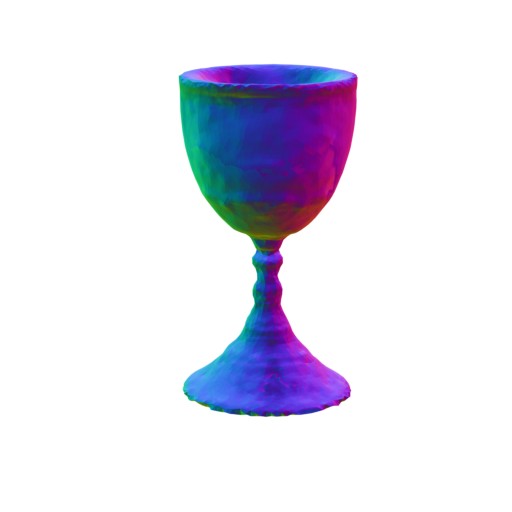} &
\includegraphics[width=3cm]{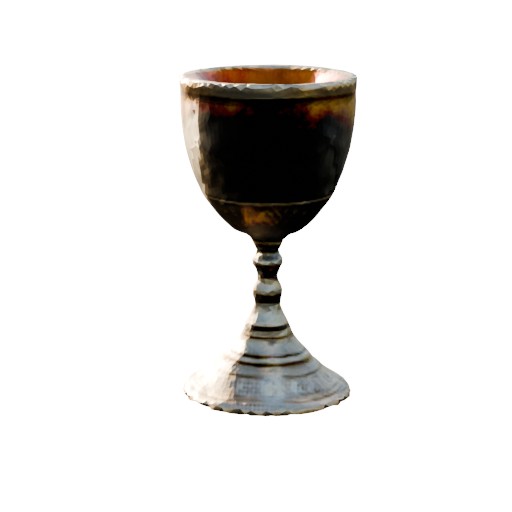} &
\includegraphics[width=3cm]{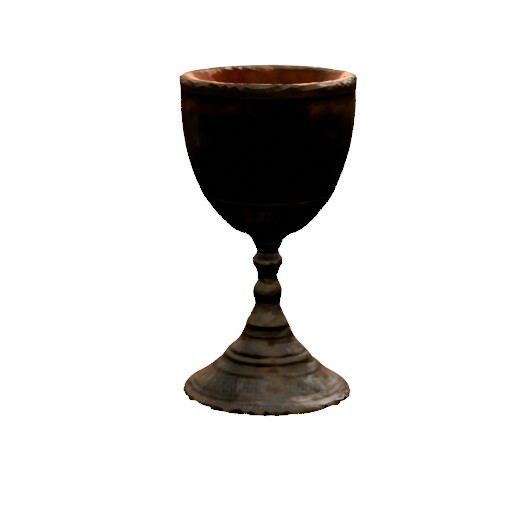} \\[-0.25cm]
\midrule

\rotatebox[origin=c]{90}{GT} & 
\includegraphics[width=3cm]{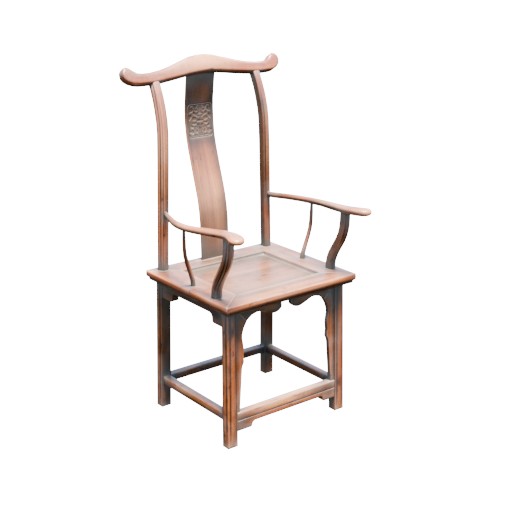} &
\includegraphics[width=3cm]{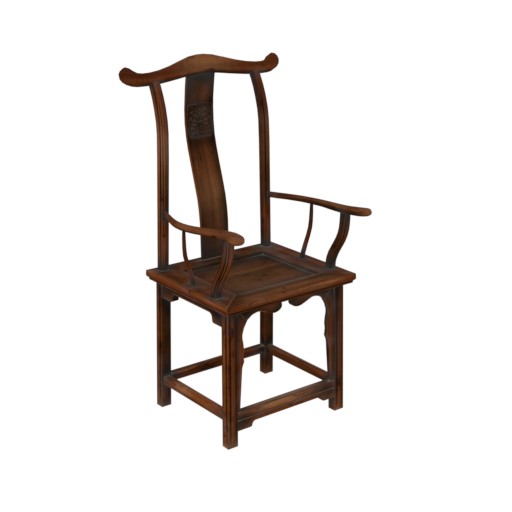} &
\includegraphics[width=3cm]{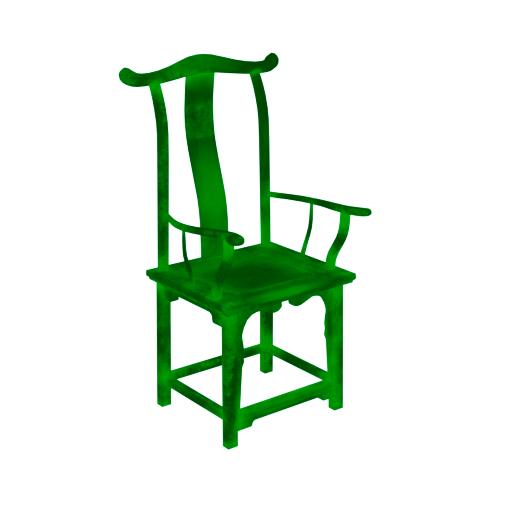} &
\includegraphics[width=3cm]{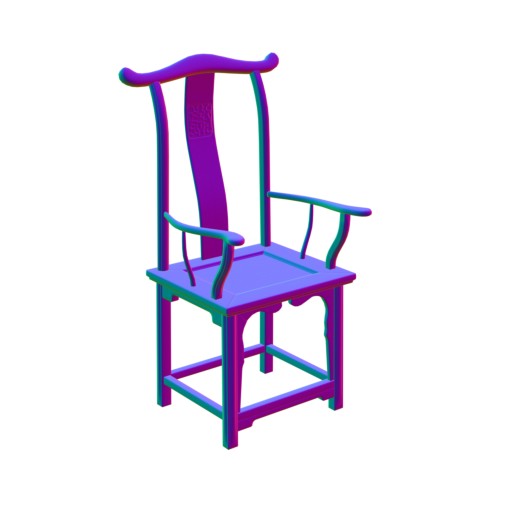} &
\includegraphics[width=3cm]{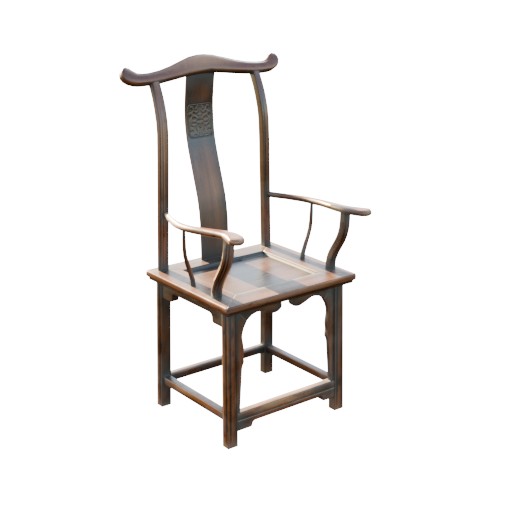} &
\includegraphics[width=3cm]{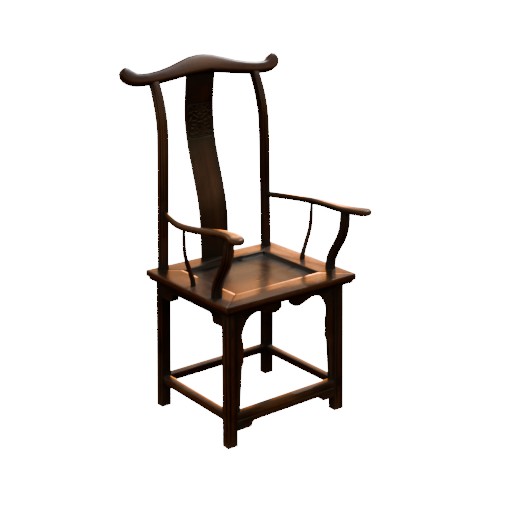} \\[-0.45cm]

\rotatebox[origin=c]{90}{Ours} & 
\includegraphics[width=3cm]{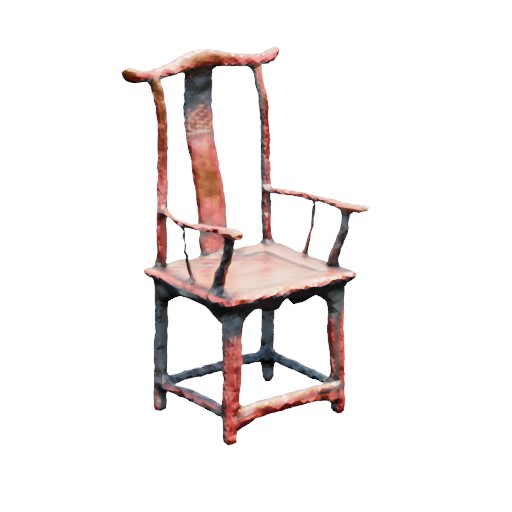} &
\includegraphics[width=3cm]{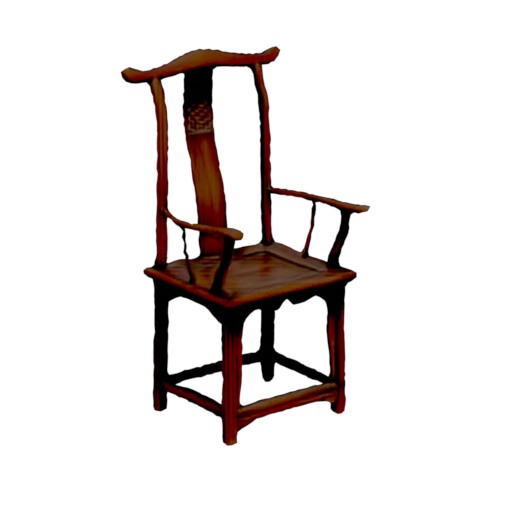} &
\includegraphics[width=3cm]{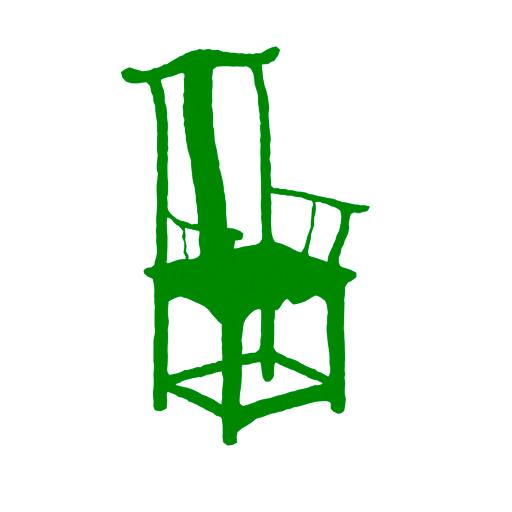} &
\includegraphics[width=3cm]{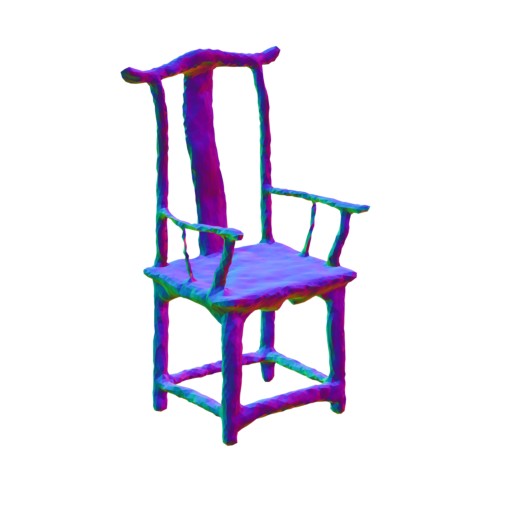} &
\includegraphics[width=3cm]{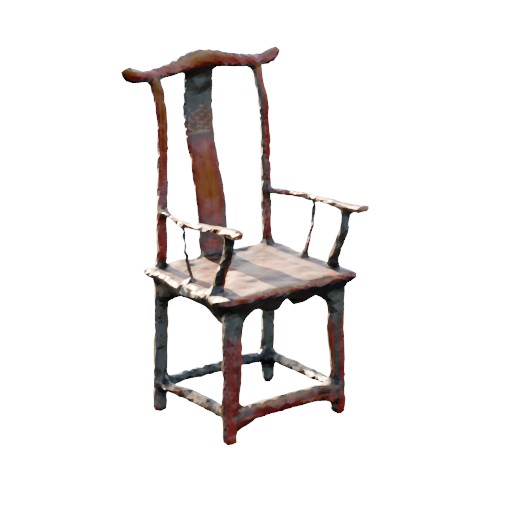} &
\includegraphics[width=3cm]{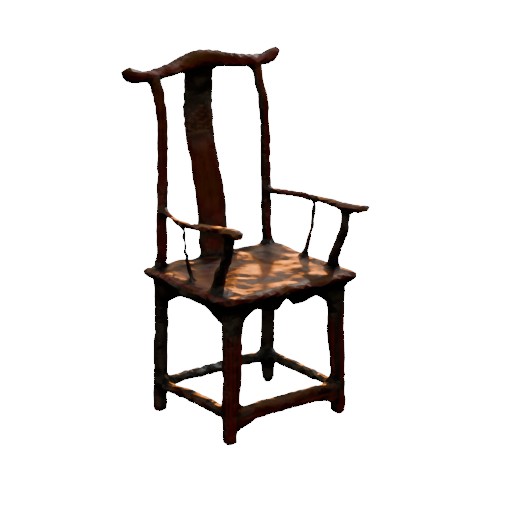} \\[-0.2cm]

\end{tabular}
} 
\vspace{-2mm}
\titlecaptionof{figure}{Decomposition Results}{%
Here, we use high-quality objects from Polyhaven~\cite{polyhaven} and render them under natural illumination. These illuminations are highly challenging for material estimation. Still our model estimates sensible material properties, which allow for a convincing relighting.
}
    \label{fig:decomposition_results}
\end{figure}

\begin{figure}[ht!]
\centering
\begin{tikzpicture}
\tikzstyle{every node}=[font=\small]
\begin{axis}[
        enlargelimits=0.05,
        legend style={at={(0.5,-0.35)},
                anchor=north,legend columns=4,
                /tikz/every even column/.append style={column sep=0.25cm}},
        xlabel={Inference time image/seconds \textdownarrow}, 
        xmode=log,
        ylabel={F-Score \textuparrow}, 
        yticklabel style={
                /pgf/number format/fixed,
                 /pgf/number format/precision=3
        },
        scaled y ticks=false,
        width=0.9\linewidth,
        height=0.65\linewidth,
        xtick pos=left,
        ytick pos=left,
        tick align=outside,
        xlabel near ticks,
        ylabel near ticks,
        axis on top,
        every x tick/.style={color=black, thin},
        every y tick/.style={color=black, thin},
        minor x tick num=1,
        grid=both,
        grid style={line width=0pt, draw=gray!0},
        major grid style={line width=.3pt,draw=gray!50},
        scatter/classes={SF3D={mark=triangle*,draw=black,fill=green!50, mark size=3pt}, TripoSR={mark=*,draw=red,fill=red!50, mark size=2.5pt}, OpenLRM={mark=*,draw=blue,fill=blue!50, mark size=2.5pt}, LGM={mark=*,draw=cyan,fill=cyan!50, mark size=2.5pt}, CRM={mark=*,draw=magenta,fill=magenta!50, mark size=2.5pt}, InstantMesh={mark=*,draw=black,fill=yellow!50, mark size=2.5pt}, ZeroShape={mark=*,draw=red,fill=black!50, mark size=2.5pt}}
]
\addplot[scatter,only marks,%
mark=o,mark size=2pt,scatter src=explicit symbolic]
table[meta=label] {
x y label
2.0827999999999998 0.4962 OpenLRM
0.31055 0.6589499999999999 TripoSR
0.47045000000000003 0.7132000000000001 SF3D
64.6152 0.3598 LGM
10.28215 0.44045 CRM
32.387950000000004 0.5549 InstantMesh
0.9545 0.49815 ZeroShape
};
\legend{SF3D, TripoSR, OpenLRM, LGM, CRM, InstantMesh, ZeroShape}
\end{axis}
\end{tikzpicture}
\titlecaption{Image-To-Mesh Time \vs Reconstruction Quality}{Our method is not only one of the fastest reconstruction methods, it is also capable of producing highly accurate geometry.}
\label{fig:speed_vs_quality}
\end{figure}

\begin{table}[ht!]
\centering
\resizebox{0.95\linewidth}{!}{ %
    \begin{tabular}{lcccc}
           & \multicolumn{4}{c}{GSO} \\%
           \cmidrule(lr){2-5} 
        Method & CD\textdownarrow & FS$@$0.1\textuparrow & FS$@$0.2\textuparrow & FS$@$0.5\textuparrow \\
        \midrule
        TripoSR~\cite{tochilkin2024triposr}  & 0.111 & 0.645 & 0.869 & 0.980 \\
        SF3D w/o Enh. Transformer  & \secondbest{0.108} & \secondbest{0.660} & \secondbest{0.872} & \secondbest{0.982} \\
        \midrule
        SF3D (Ours)            & \best{0.098} & \best{0.701} & \best{0.894} & \best{0.988}  
    \end{tabular}
}
\titlecaption{Ablation}{
As our model builds upon TripoSR, we use it as our baseline.
Without our enhanced transformer, our mesh training has already improved upon it. With our architectural improvements, we outperform the baselines significantly.
} 
\label{tab:ablation}
\end{table}

\inlinesection{Evaluations.}
For runtime comparisons, we consider the meshes as the final output and calculate the entire run time required to go from the input image to the final mesh.
We then perform separate evaluations for the shape quality.
Several models cannot be conditioned on intrinsic and extrinsic camera parameters, so we propose performing an alignment step.
We normalize the mesh, perform brute-force rotation alignment tests, and select the rotation with the lowest Chamfer Distance (CD). 
We then run another alignment step using Iterative Closest Point (ICP), where we further optimize the rotation and translation.
We then calculate the standard shape metrics of CD and F-score (FS) after this alignment.
This alignment might still misalign symmetric objects for re-rendering. Hence, we only report indicative rendering metrics in the supplements.

\inlinesection{Triangle Counts.}
We run each method under the default configuration. As the triangle count varies drastically based on the 3D mesh, we report the triangle count for a single mesh here: InstantMesh 57.3K, CRM 24.1K, LGM 42.1K, TripoSR 32.1K, OpenLRM 662K, Ours 27.4K. 

\inlinesection{Results.}
In \Table{comparison_3d}, we compare our method with all the baselines. 
Here, our method outperforms all current and concurrent baselines on both CD and F-scores. This indicates that our method can reconstruct accurate shapes, even if our models have fewer polygons than the other baselines.
From the visual comparisons in~\fig{multi_comparison_gso_omniobj}, it can be seen that the accurate shape reconstruction of SF3D also translates well to the visual quality of the 3D assets.
Here, it is worth noting that SF3D also handles fine geometry like the glasses well and produces consistent shapes with more detailed textures than the SOTA methods.
Moreover, the results also show sensible material properties and albedo as seen in \fig{decomposition_results}.
This is highly challenging as the objects are only rendered under natural illumination. 
Estimating material properties without any knowledge about the illumination is a highly ambiguous problem.

\inlinesection{Inference Speed \vs Reconstruction Quality.}
In \fig{speed_vs_quality}, we plot the inference speed \vs reconstruction quality for different techniques. 
The best-performing methods should be located in the top left corner. 
While our method is slightly slower than TripoSR, the reconstruction accuracy is considerably better for SF3D.
It is also worth noting that our reconstruction has smoother shading with less pronounced marching cube artifacts, as seen in \fig{multi_comparison_gso_omniobj}.
The final apparent asset quality is thus even higher for our method.

\inlinesection{Ablations.}
We evaluate our additions in \Table{ablation} against baseline models.
Our method is built upon TripoSR, so we use it as our initial baseline.
If we add mesh training and relighting as a fine-tuning step, we can see that 'SF3D w/o Enhanced Transformer' already improves upon TripoSR, demonstrating the use of our mesh-based training. 
This improvement is mainly due to the efficient rendering, enabling higher resolution supervision during training and smoother meshes from vertex offsets.
If we add our high-resolution triplanes using our enhanced transformer, SF3D further outperforms the baseline considerably.

\inlinesection{Limitations and Outlook.}
As seen in \fig{multi_comparison_gso_omniobj} (top row), the cup's albedo is not perfectly matched. This is related to the LDR input, where the dark spots do not contain any useful information. 
Furthermore, our roughness and metallic properties are homogeneous, which limits their usefulness for objects containing multiple drastically different materials that are spatially-varying. 
In addition, our method also introduces material prediction and delighting without explicit supervision. As we do not require explicit supervision of these parameters, our method could be extended to train on real-world datasets, which we leave for future work. Similarly, the UV unwrapping could leverage existing datasets for further improvements.

\vspace{-2mm}
\section{Conclusion}
\vspace{-1mm}

We present SF3D, a fast single view to uv-unwrapped textured object reconstruction method. 
In addition to our fast extraction pipeline, we introduce several architectural improvements to feed-forward-based 3D reconstruction methods, which help our model produce highly detailed geometry and texture.
In our extensive evaluation, we show that our method outperforms existing and concurrent baselines in both speed and quality.
{
    \small
    \bibliographystyle{ieeenat_fullname}
    \bibliography{references}

\begin{thebibliography}{82}
\providecommand{\natexlab}[1]{#1}
\providecommand{\url}[1]{\texttt{#1}}
\expandafter\ifx\csname urlstyle\endcsname\relax
  \providecommand{\doi}[1]{doi: #1}\else
  \providecommand{\doi}{doi: \begingroup \urlstyle{rm}\Url}\fi

\bibitem[Adelson and Pentland(1996)]{Adelson1996}
E.H. Adelson and A.P. Pentland.
\newblock \emph{The perception of shading and reflectance}, page 409–424.
\newblock Cambridge University Press, 1996.

\bibitem[Blattmann et~al.(2023{\natexlab{a}})Blattmann, Dockhorn, Kulal, Mendelevitch, Kilian, Lorenz, Levi, English, Voleti, Letts, et~al.]{blattmann2023stable}
Andreas Blattmann, Tim Dockhorn, Sumith Kulal, Daniel Mendelevitch, Maciej Kilian, Dominik Lorenz, Yam Levi, Zion English, Vikram Voleti, Adam Letts, et~al.
\newblock Stable {V}ideo {D}iffusion: Scaling latent video diffusion models to large datasets.
\newblock \emph{arXiv}, 2023{\natexlab{a}}.

\bibitem[Blattmann et~al.(2023{\natexlab{b}})Blattmann, Rombach, Ling, Dockhorn, Kim, Fidler, and Kreis]{blattmann2023align}
Andreas Blattmann, Robin Rombach, Huan Ling, Tim Dockhorn, Seung~Wook Kim, Sanja Fidler, and Karsten Kreis.
\newblock {Align your Latents: High-Resolution Video Synthesis with Latent Diffusion Models}.
\newblock \emph{arXiv}, 2023{\natexlab{b}}.

\bibitem[Boss et~al.(2021{\natexlab{a}})Boss, Braun, Jampani, Barron, Liu, and Lensch]{Boss2021}
Mark Boss, Raphael Braun, Varun Jampani, Jonathan~T. Barron, Ce Liu, and Hendrik~P.A. Lensch.
\newblock {NeRD}: Neural reflectance decomposition from image collections.
\newblock \emph{ICCV}, 2021{\natexlab{a}}.

\bibitem[Boss et~al.(2021{\natexlab{b}})Boss, Jampani, Braun, Liu, Barron, and Lensch]{Boss2021neuralPIL}
Mark Boss, Varun Jampani, Raphael Braun, Ce Liu, Jonathan~T. Barron, and Hendrik~P.A. Lensch.
\newblock Neural-pil: Neural pre-integrated lighting for reflectance decomposition.
\newblock \emph{NeurIPS}, 2021{\natexlab{b}}.

\bibitem[Boss et~al.(2022)Boss, Engelhardt, Kar, Li, Sun, Barron, Lensch, and Jampani]{bossSAMURAIShapeMaterial2022}
Mark Boss, Andreas Engelhardt, Abhishek Kar, Yuanzhen Li, Deqing Sun, Jonathan~T. Barron, Hendrik~P.A. Lensch, and Varun Jampani.
\newblock {{SAMURAI}}: {{Shape And Material}} from {{Unconstrained Real-world Arbitrary Image}} collections.
\newblock \emph{NeurIPS}, 2022.

\bibitem[Burley(2012)]{Burley2012}
Brent Burley.
\newblock Physically based shading at disney.
\newblock \emph{ACM Transactions on Graphics (SIGGRAPH)}, 2012.

\bibitem[Caron et~al.(2021)Caron, Touvron, Misra, J\'egou, Mairal, Bojanowski, and Joulin]{caron2021emerging}
Mathilde Caron, Hugo Touvron, Ishan Misra, Herv\'e J\'egou, Julien Mairal, Piotr Bojanowski, and Armand Joulin.
\newblock Emerging properties in self-supervised vision transformers.
\newblock \emph{ICCV}, 2021.

\bibitem[Chan et~al.(2021)Chan, Lin, Chan, Nagano, Pan, Mello, Gallo, Guibas, Tremblay, Khamis, Karras, and Wetzstein]{Chan2021eg3d}
Eric~R. Chan, Connor~Z. Lin, Matthew~A. Chan, Koki Nagano, Boxiao Pan, Shalini~De Mello, Orazio Gallo, Leonidas Guibas, Jonathan Tremblay, Sameh Khamis, Tero Karras, and Gordon Wetzstein.
\newblock Efficient geometry-aware {3D} generative adversarial networks.
\newblock In \emph{arXiv}, 2021.

\bibitem[Chen et~al.(2023)Chen, Chen, Jiao, and Jia]{chen2023fantasia3d}
Rui Chen, Yongwei Chen, Ningxin Jiao, and Kui Jia.
\newblock Fantasia{3D}: Disentangling geometry and appearance for high-quality text-to-{3D} content creation.
\newblock In \emph{ICCV}, 2023.

\bibitem[Deitke et~al.(2023)Deitke, Liu, Wallingford, Ngo, Michel, Kusupati, Fan, Laforte, Voleti, Gadre, et~al.]{deitke2023objaversexl}
Matt Deitke, Ruoshi Liu, Matthew Wallingford, Huong Ngo, Oscar Michel, Aditya Kusupati, Alan Fan, Christian Laforte, Vikram Voleti, Samir~Yitzhak Gadre, et~al.
\newblock Objaverse-{XL}: A universe of 10m+ 3{D} objects.
\newblock \emph{arXiv}, 2023.

\bibitem[Downs et~al.(2022)Downs, Francis, Koenig, Kinman, Hickman, Reymann, McHugh, and Vanhoucke]{downs2022gso}
Laura Downs, Anthony Francis, Nate Koenig, Brandon Kinman, Ryan Hickman, Krista Reymann, Thomas~B McHugh, and Vincent Vanhoucke.
\newblock Google {S}canned {O}bjects: A high-quality dataset of 3{D} scanned household items.
\newblock In \emph{2022 International Conference on Robotics and Automation (ICRA)}, pages 2553--2560. IEEE, 2022.

\bibitem[Engelhardt et~al.(2024)Engelhardt, Raj, Boss, Zhang, Kar, Li, Sun, Barron, Lensch, and Jampani]{engelhardt2023-shinobi}
Andreas Engelhardt, Amit Raj, Mark Boss, Yunzhi Zhang, Abhishek Kar, Yuanzhen Li, Deqing Sun, Jonathan~T. Barron, Hendrik~P.A. Lensch, and Varun Jampani.
\newblock {SHINOBI}: {Sh}ape and {I}llumination using {N}eural {O}bject decomposition via {B}rdf optimization {I}n-the-wild.
\newblock In \emph{CVPR}, 2024.

\bibitem[Gao et~al.(2024)Gao, Holynski, Henzler, Brussee, Martin-Brualla, Srinivasan, Barron, and Poole]{Gao2024CAT3DCA}
Ruiqi Gao, Aleksander Holynski, Philipp Henzler, Arthur Brussee, Ricardo Martin-Brualla, Pratul~P. Srinivasan, Jonathan~T. Barron, and Ben Poole.
\newblock {CAT3D}: Create anything in {3D} with multi-view diffusion models.
\newblock \emph{arXiv}, 2024.

\bibitem[Girdhar et~al.(2023)Girdhar, Singh, Brown, Duval, Azadi, Rambhatla, Shah, Yin, Parikh, and Misra]{Rohit2023EMU}
Rohit Girdhar, Mannat Singh, Andrew Brown, Quentin Duval, Samaneh Azadi, Sai~Saketh Rambhatla, Akbar Shah, Xi Yin, Devi Parikh, and Ishan Misra.
\newblock {EMU VIDEO: Factorizing Text-to-Video Generation by Explicit Image Conditioning}, 2023.

\bibitem[Guo et~al.(2023)Guo, Liu, Shao, Laforte, Voleti, Luo, Chen, Zou, Wang, Cao, and Zhang]{threestudio2023}
Yuan-Chen Guo, Ying-Tian Liu, Ruizhi Shao, Christian Laforte, Vikram Voleti, Guan Luo, Chia-Hao Chen, Zi-Xin Zou, Chen Wang, Yan-Pei Cao, and Song-Hai Zhang.
\newblock threestudio: A unified framework for 3d content generation.
\newblock \url{https://github.com/threestudio-project/threestudio}, 2023.

\bibitem[Hasselgren et~al.(20222)Hasselgren, Hofmann, and Munkberg]{hasselgrenShapeLightMaterial2022}
Jon Hasselgren, Nikolai Hofmann, and Jacob Munkberg.
\newblock Shape, {{Light}} \& {{Material Decomposition}} from {{Images}} using {{Monte Carlo Rendering}} and {{Denoising}}.
\newblock \emph{NeurIPS}, 20222.

\bibitem[He and Wang(2023)]{openlrm}
Zexin He and Tengfei Wang.
\newblock {OpenLRM}: Open-source large reconstruction models.
\newblock \url{https://github.com/3DTopia/OpenLRM}, 2023.

\bibitem[Ho et~al.(2020)Ho, Jain, and Abbeel]{ho2020ddpm}
Jonathan Ho, Ajay Jain, and Pieter Abbeel.
\newblock Denoising diffusion probabilistic models.
\newblock In \emph{NeurIPS}, 2020.

\bibitem[Hong et~al.(2024)Hong, Zhang, Gu, Bi, Zhou, Liu, Liu, Sunkavalli, Bui, and Tan]{hong2024lrm}
Yicong Hong, Kai Zhang, Jiuxiang Gu, Sai Bi, Yang Zhou, Difan Liu, Feng Liu, Kalyan Sunkavalli, Trung Bui, and Hao Tan.
\newblock {LRM}: Large reconstruction model for single image to {3D}.
\newblock \emph{ICLR}, 2024.

\bibitem[Huang et~al.(2023)Huang, Stojanov, Thai, Jampani, and Rehg]{Huang2023ZeroShapeRZ}
Zixuan Huang, Stefan Stojanov, Anh Thai, Varun Jampani, and James~M. Rehg.
\newblock {ZeroShape}: Regression-based zero-shot shape reconstruction.
\newblock \emph{arXiv}, 2023.

\bibitem[Huang et~al.(2024)Huang, Johnson, Debnath, Rehg, and Wu]{huang2024pointinfinity}
Zixuan Huang, Justin Johnson, Shoubhik Debnath, James~M Rehg, and Chao-Yuan Wu.
\newblock Pointinfinity: Resolution-invariant point diffusion models.
\newblock In \emph{CVPR}, 2024.

\bibitem[Jiang et~al.(2024)Jiang, Huang, and Pavlakos]{Jiang2024Real3DSU}
Hanwen Jiang, Qixing Huang, and Georgios Pavlakos.
\newblock {Real3D}: Scaling up large reconstruction models with real-world images.
\newblock \emph{arXiv}, 2024.

\bibitem[Kerbl et~al.(2023)Kerbl, Kopanas, Leimk{\"u}hler, and Drettakis]{kerbl3Dgaussians}
Bernhard Kerbl, Georgios Kopanas, Thomas Leimk{\"u}hler, and George Drettakis.
\newblock 3d gaussian splatting for real-time radiance field rendering.
\newblock \emph{ACM TOG}, 42\penalty0 (4), 2023.

\bibitem[Kong et~al.(2024)Kong, Liu, Lyu, Taher, Qi, and Davison]{kong2024eschernet}
Xin Kong, Shikun Liu, Xiaoyang Lyu, Marwan Taher, Xiaojuan Qi, and Andrew~J Davison.
\newblock {EscherNet}: A generative model for scalable view synthesis.
\newblock \emph{arXiv}, 2024.

\bibitem[Kwak et~al.(2024)Kwak, Dong, Jin, Ko, Mahajan, and Yi]{kwak2023vivid}
Jeong-gi Kwak, Erqun Dong, Yuhe Jin, Hanseok Ko, Shweta Mahajan, and Kwang~Moo Yi.
\newblock {ViVid-1-to-3}: Novel view synthesis with video diffusion models.
\newblock \emph{CVPR}, 2024.

\bibitem[Levy(2024)]{geogram}
Bruno Levy.
\newblock geogram.
\newblock \url{https://github.com/BrunoLevy/geogram}, 2024.

\bibitem[Li et~al.(2023)Li, Tan, Zhang, Xu, Luan, Xu, Hong, Sunkavalli, Shakhnarovich, and Bi]{Li2023Instant3DFT}
Jiahao Li, Hao Tan, Kai Zhang, Zexiang Xu, Fujun Luan, Yinghao Xu, Yicong Hong, Kalyan Sunkavalli, Greg Shakhnarovich, and Sai Bi.
\newblock {Instant3D}: Fast text-to-{3D} with sparse-view generation and large reconstruction model.
\newblock \emph{arXiv}, 2023.

\bibitem[Liu et~al.(2023{\natexlab{a}})Liu, Shi, Chen, Zhang, Xu, Wei, Chen, Zeng, Gu, and Su]{liu2023one2345pp}
Minghua Liu, Ruoxi Shi, Linghao Chen, Zhuoyang Zhang, Chao Xu, Xinyue Wei, Hansheng Chen, Chong Zeng, Jiayuan Gu, and Hao Su.
\newblock One-2-3-45++: Fast single image to {3D} objects with consistent multi-view generation and {3D} diffusion.
\newblock \emph{arXiv}, 2023{\natexlab{a}}.

\bibitem[Liu et~al.(2023{\natexlab{b}})Liu, Xu, Jin, Chen, Varma~T, Xu, and Su]{liu2023one2345}
Minghua Liu, Chao Xu, Haian Jin, Linghao Chen, Mukund Varma~T, Zexiang Xu, and Hao Su.
\newblock One-2-3-45: Any single image to 3{D} mesh in 45 seconds without per-shape optimization.
\newblock \emph{NeurIPS}, 2023{\natexlab{b}}.

\bibitem[Liu et~al.(2023{\natexlab{c}})Liu, Wu, Hoorick, Tokmakov, Zakharov, and Vondrick]{liu2023zero1to3}
Ruoshi Liu, Rundi Wu, Basile~Van Hoorick, Pavel Tokmakov, Sergey Zakharov, and Carl Vondrick.
\newblock Zero-1-to-3: Zero-shot one image to 3{D} object.
\newblock \emph{ICCV}, 2023{\natexlab{c}}.

\bibitem[Liu et~al.(2023{\natexlab{d}})Liu, Lin, Zeng, Long, Liu, Komura, and Wang]{liu2023syncdreamer}
Yuan Liu, Cheng Lin, Zijiao Zeng, Xiaoxiao Long, Lingjie Liu, Taku Komura, and Wenping Wang.
\newblock {SyncDreamer}: Generating multiview-consistent images from a single-view image.
\newblock \emph{arXiv}, 2023{\natexlab{d}}.

\bibitem[Liu et~al.(2023{\natexlab{e}})Liu, Li, Lin, Yu, Peng, Cao, Qi, Huang, Liang, and Ouyang]{Liu2023UniDreamUD}
Zexiang Liu, Yangguang Li, Youtian Lin, Xin Yu, Sida Peng, Yan-Pei Cao, Xiaojuan Qi, Xiaoshui Huang, Ding Liang, and Wanli Ouyang.
\newblock Unidream: Unifying diffusion priors for relightable text-to-{3D} generation.
\newblock \emph{arXiv}, 2023{\natexlab{e}}.

\bibitem[Long et~al.(2023)Long, Guo, Lin, Liu, Dou, Liu, Ma, Zhang, Habermann, Theobalt, et~al.]{long2023wonder3d}
Xiaoxiao Long, Yuan-Chen Guo, Cheng Lin, Yuan Liu, Zhiyang Dou, Lingjie Liu, Yuexin Ma, Song-Hai Zhang, Marc Habermann, Christian Theobalt, et~al.
\newblock {Wonder3D}: Single image to {3D} using cross-domain diffusion.
\newblock \emph{arXiv}, 2023.

\bibitem[Lorensen and Cline(1987)]{lorensen1987mc}
William~E. Lorensen and Harvey~E. Cline.
\newblock Marching cubes: A high resolution 3d surface construction algorithm.
\newblock \emph{ACM Transactions on Graphics (SIGGRAPH)}, 1987.

\bibitem[Melas-Kyriazi et~al.(2024)Melas-Kyriazi, Laina, Rupprecht, Neverova, Vedaldi, Gafni, and Kokkinos]{melas2024im3d}
Luke Melas-Kyriazi, Iro Laina, Christian Rupprecht, Natalia Neverova, Andrea Vedaldi, Oran Gafni, and Filippos Kokkinos.
\newblock {IM-3D}: Iterative multiview diffusion and reconstruction for high-quality {3D} generation.
\newblock \emph{arXiv}, 2024.

\bibitem[Mercier et~al.(2024)Mercier, Nakhli, Reddy, Yasarla, Cai, Porikli, and Berger]{mercier2024hexagen3d}
Antoine Mercier, Ramin Nakhli, Mahesh Reddy, Rajeev Yasarla, Hong Cai, Fatih Porikli, and Guillaume Berger.
\newblock {HexaGen3D}: Stablediffusion is just one step away from fast and diverse {Text-to-3D} generation.
\newblock \emph{arXiv}, 2024.

\bibitem[Mildenhall et~al.(2020)Mildenhall, Srinivasan, Tancik, Barron, Ramamoorthi, and Ng]{mildenhall2020}
Ben Mildenhall, Pratul Srinivasan, Matthew Tancik, Jonathan~T. Barron, Ravi Ramamoorthi, and Ren Ng.
\newblock {NeRF}: Representing scenes as neural radiance fields for view synthesis.
\newblock \emph{ECCV}, 2020.

\bibitem[Munkberg et~al.(2022)Munkberg, Hasselgren, Shen, Gao, Chen, Evans, Mueller, and Fidler]{munkberg2022nvdiffrec}
Jacob Munkberg, Jon Hasselgren, Tianchang Shen, Jun Gao, Wenzheng Chen, Alex Evans, Thomas Mueller, and Sanja Fidler.
\newblock {Extracting Triangular 3D Models, Materials, and Lighting From Images}.
\newblock \emph{CVPR}, 2022.

\bibitem[Oquab et~al.(2023)Oquab, Darcet, Moutakanni, Vo, Szafraniec, Khalidov, Fernandez, Haziza, Massa, El-Nouby, Howes, Huang, Xu, Sharma, Li, Galuba, Rabbat, Assran, Ballas, Synnaeve, Misra, Jegou, Mairal, Labatut, Joulin, and Bojanowski]{oquab2023dinov2}
Maxime Oquab, Timothée Darcet, Theo Moutakanni, Huy~V. Vo, Marc Szafraniec, Vasil Khalidov, Pierre Fernandez, Daniel Haziza, Francisco Massa, Alaaeldin El-Nouby, Russell Howes, Po-Yao Huang, Hu Xu, Vasu Sharma, Shang-Wen Li, Wojciech Galuba, Mike Rabbat, Mido Assran, Nicolas Ballas, Gabriel Synnaeve, Ishan Misra, Herve Jegou, Julien Mairal, Patrick Labatut, Armand Joulin, and Piotr Bojanowski.
\newblock Dinov2: Learning robust visual features without supervision, 2023.

\bibitem[Poole et~al.(2022)Poole, Jain, Barron, and Mildenhall]{poole2022dreamfusion}
Ben Poole, Ajay Jain, Jonathan~T. Barron, and Ben Mildenhall.
\newblock Dreamfusion: Text-to-{3D} using 2d diffusion.
\newblock \emph{arXiv}, 2022.

\bibitem[Radford et~al.(2021)Radford, Kim, Hallacy, Ramesh, Goh, Agarwal, Sastry, Askell, Mishkin, Clark, et~al.]{radford2021clip}
Alec Radford, Jong~Wook Kim, Chris Hallacy, Aditya Ramesh, Gabriel Goh, Sandhini Agarwal, Girish Sastry, Amanda Askell, Pamela Mishkin, Jack Clark, et~al.
\newblock Learning transferable visual models from natural language supervision.
\newblock In \emph{International conference on machine learning}, pages 8748--8763. PMLR, 2021.

\bibitem[Rombach et~al.(2022)Rombach, Blattmann, Lorenz, Esser, and Ommer]{rombach2022high}
Robin Rombach, Andreas Blattmann, Dominik Lorenz, Patrick Esser, and Bj{\"o}rn Ommer.
\newblock High-resolution image synthesis with latent diffusion models.
\newblock In \emph{CVPR}, pages 10684--10695, 2022.

\bibitem[Ruiz et~al.(2022)Ruiz, Li, Jampani, Pritch, Rubinstein, and Aberman]{ruiz2022dreambooth}
Nataniel Ruiz, Yuanzhen Li, Varun Jampani, Yael Pritch, Michael Rubinstein, and Kfir Aberman.
\newblock {DreamBooth}: Fine tuning text-to-image dissusion models for subject-driven generation.
\newblock \emph{arXiv}, 2022.

\bibitem[Sauer et~al.(2023)Sauer, Lorenz, Blattmann, and Rombach]{sauer2023add}
Axel Sauer, Dominik Lorenz, Andreas Blattmann, and Robin Rombach.
\newblock Adversarial diffusion distillation.
\newblock \emph{arXiv}, 2023.

\bibitem[Shen et~al.(2021)Shen, Gao, Yin, Liu, and Fidler]{shen2021dmtet}
Tianchang Shen, Jun Gao, Kangxue Yin, Ming-Yu Liu, and Sanja Fidler.
\newblock {D}eep {M}arching {T}etrahedra: a hybrid representation for high-resolution 3{D} shape synthesis.
\newblock In \emph{Advances in Neural Information Processing Systems (NeurIPS)}, 2021.

\bibitem[Shi et~al.(2023{\natexlab{a}})Shi, Chen, Zhang, Liu, Xu, Wei, Chen, Zeng, and Su]{shi2023zero123pp}
Ruoxi Shi, Hansheng Chen, Zhuoyang Zhang, Minghua Liu, Chao Xu, Xinyue Wei, Linghao Chen, Chong Zeng, and Hao Su.
\newblock Zero123++: a single image to consistent multi-view diffusion base model.
\newblock \emph{arXiv}, 2023{\natexlab{a}}.

\bibitem[Shi et~al.(2016)Shi, Caballero, Huszár, Totz, Aitken, Bishop, Rueckert, and Wang]{shi2016realtime}
Wenzhe Shi, Jose Caballero, Ferenc Huszár, Johannes Totz, Andrew~P. Aitken, Rob Bishop, Daniel Rueckert, and Zehan Wang.
\newblock Real-time single image and video super-resolution using an efficient sub-pixel convolutional neural network.
\newblock \emph{CVPR}, 2016.

\bibitem[Shi et~al.(2023{\natexlab{b}})Shi, Wang, Ye, Long, Li, and Yang]{shi2023mvdream}
Yichun Shi, Peng Wang, Jianglong Ye, Mai Long, Kejie Li, and Xiao Yang.
\newblock {MVDream}: Multi-view diffusion for 3d generation.
\newblock \emph{arXiv}, 2023{\natexlab{b}}.

\bibitem[Song et~al.(2020)Song, Sohl-Dickstein, Kingma, Kumar, Ermon, and Poole]{song2020sgm}
Yang Song, Jascha Sohl-Dickstein, Diederik~P Kingma, Abhishek Kumar, Stefano Ermon, and Ben Poole.
\newblock Score-based generative modeling through stochastic differential equations.
\newblock \emph{arXiv}, 2020.

\bibitem[StabilityAI(2023)]{stablezero123}
StabilityAI.
\newblock {Stable Zero123}, 2023.

\bibitem[Szymanowicz et~al.(2024)Szymanowicz, Rupprecht, and Vedaldi]{Szymanowicz2024SplatterIU}
Stanislaw Szymanowicz, C. Rupprecht, and Andrea Vedaldi.
\newblock Splatter image: Ultra-fast single-view {3D} reconstruction.
\newblock \emph{CVPR}, 2024.

\bibitem[Tang et~al.(2024)Tang, Chen, Chen, Wang, Zeng, and Liu]{Tang2024LGMLM}
Jiaxiang Tang, Zhaoxi Chen, Xiaokang Chen, Tengfei Wang, Gang Zeng, and Ziwei Liu.
\newblock {LGM}: Large multi-view gaussian model for high-resolution {3D} content creation.
\newblock \emph{arXiv}, 2024.

\bibitem[Tochilkin et~al.(2024)Tochilkin, Pankratz, Liu, Huang, Letts, Li, Liang, Laforte, Jampani, and Cao]{tochilkin2024triposr}
Dmitry Tochilkin, David Pankratz, Zexiang Liu, Zixuan Huang, Adam Letts, Yangguang Li, Ding Liang, Christian Laforte, Varun Jampani, and Yan-Pei Cao.
\newblock {TripoSR}: Fast {3D} object reconstruction from a single image.
\newblock \emph{arXiv}, 2024.

\bibitem[Vainer et~al.(2024)Vainer, Boss, Parger, Kutsy, De~Nigris, Rowles, Perony, and Donné]{Vainer2024CollaborativeCF}
Shimon Vainer, Mark Boss, Mathias Parger, Konstantin Kutsy, Dante De~Nigris, Ciara Rowles, Nicolas Perony, and Simon Donné.
\newblock Collaborative control for geometry-conditioned {PBR} image generation.
\newblock \emph{arXiv}, 2024.

\bibitem[Voleti et~al.(2022)Voleti, Jolicoeur-Martineau, and Pal]{voleti2022mcvd}
Vikram Voleti, Alexia Jolicoeur-Martineau, and Christopher Pal.
\newblock {MCVD}: Masked conditional video diffusion for prediction, generation, and interpolation.
\newblock In \emph{NeurIPS}, 2022.

\bibitem[Voleti et~al.(2024)Voleti, Yao, Boss, Letts, Pankratz, Tochilkin, Laforte, Rombach, and Jampani]{voleti2024sv3d}
Vikram Voleti, Chun-Han Yao, Mark Boss, Adam Letts, David Pankratz, Dmitrii Tochilkin, Christian Laforte, Robin Rombach, and Varun Jampani.
\newblock {SV3D}: Novel multi-view synthesis and {3D} generation from a single image using latent video diffusion.
\newblock \emph{arXiv}, 2024.

\bibitem[Wang et~al.(2023)Wang, Tan, Bi, Xu, Luan, Sunkavalli, Wang, Xu, and Zhang]{Wang2023PFLRMPL}
Peng Wang, Hao Tan, Sai Bi, Yinghao Xu, Fujun Luan, Kalyan Sunkavalli, Wenping Wang, Zexiang Xu, and Kai Zhang.
\newblock {PF-LRM}: Pose-free large reconstruction model for joint pose and shape prediction.
\newblock \emph{arXiv}, 2023.

\bibitem[Wang et~al.(2024)Wang, Wang, Chen, Xiang, Chen, Yu, Li, Su, and Zhu]{Wang2024CRMSI}
Zhengyi Wang, Yikai Wang, Yifei Chen, Chendong Xiang, Shuo Chen, Dajiang Yu, Chongxuan Li, Hang Su, and Jun Zhu.
\newblock {CRM}: Single image to {3D} textured mesh with convolutional reconstruction model.
\newblock \emph{arXiv}, 2024.

\bibitem[Wei et~al.(2024)Wei, Zhang, Bi, Tan, Luan, Deschaintre, Sunkavalli, Su, and Xu]{Wei2024MeshLRMLR}
Xinyue Wei, Kai Zhang, Sai Bi, Hao Tan, Fujun Luan, Valentin Deschaintre, Kalyan Sunkavalli, Hao Su, and Zexiang Xu.
\newblock {MeshLRM}: Large reconstruction model for high-quality mesh.
\newblock \emph{arXiv}, 2024.

\bibitem[Wen et~al.(2024)Wen, Huang, Wang, Chen, Qiao, and Sheng]{Wen2024Ouroboros3DIG}
Hao Wen, Zehuan Huang, Yaohui Wang, Xinyuan Chen, Yu Qiao, and Lu Sheng.
\newblock {Ouroboros3D}: Image-to-{3D} generation via {3D}-aware recursive diffusion.
\newblock \emph{arXiv}, 2024.

\bibitem[Weng et~al.(2023)Weng, Yang, Wang, Li, Zhang, Chen, and Zhang]{weng2023consistent123}
Haohan Weng, Tianyu Yang, Jianan Wang, Yu Li, Tong Zhang, C.~L.~Philip Chen, and Lei Zhang.
\newblock Consistent123: Improve consistency for one image to {3D} object synthesis.
\newblock \emph{arXiv}, 2023.

\bibitem[Wu et~al.(2024)Wu, Liu, Cai, Yan, Wang, Hu, Duan, and Ma]{Wu2024Unique3DHA}
Kailu Wu, Fangfu Liu, Zhihan Cai, Runjie Yan, Hanyang Wang, Yating Hu, Yueqi Duan, and Kaisheng Ma.
\newblock {Unique3D}: High-quality and efficient {3D} mesh generation from a single image.
\newblock \emph{arXiv}, 2024.

\bibitem[Wu et~al.(2023)Wu, Zhang, Fu, Wang, Jiawei~Ren, Wu, Yang, Wang, Qian, Lin, and Liu]{wu2023omniobject3d}
Tong Wu, Jiarui Zhang, Xiao Fu, Yuxin Wang, Liang~Pan Jiawei~Ren, Wayne Wu, Lei Yang, Jiaqi Wang, Chen Qian, Dahua Lin, and Ziwei Liu.
\newblock Omniobject3d: Large-vocabulary 3d object dataset for realistic perception, reconstruction and generation.
\newblock In \emph{IEEE/CVF Conference on Computer Vision and Pattern Recognition (CVPR)}, 2023.

\bibitem[Xie et~al.(2024{\natexlab{a}})Xie, Bi, Shu, Zhang, Xu, Zhou, Pirk, Kaufman, Sun, and Tan]{Xie2024LRMZeroTL}
Desai Xie, Sai Bi, Zhixin Shu, Kai Zhang, Zexiang Xu, Yi Zhou, Soren Pirk, Arie~E. Kaufman, Xin Sun, and Hao Tan.
\newblock {LRM-Zero}: Training large reconstruction models with synthesized data.
\newblock \emph{arXiv}, 2024{\natexlab{a}}.

\bibitem[Xie et~al.(2024{\natexlab{b}})Xie, Zheng, Huang, Chen, Wang, Ye, Chen, and Huo]{Xie2024LDMLT}
Rengan Xie, Wenting Zheng, Kai Huang, Yizheng Chen, Qi Wang, Qi Ye, Wei Chen, and Yuchi Huo.
\newblock {LDM}: Large tensorial {SDF} model for textured mesh generation.
\newblock \emph{arXiv}, 2024{\natexlab{b}}.

\bibitem[Xie et~al.(2024{\natexlab{c}})Xie, Yao, Voleti, Jiang, and Jampani]{xie2024sv4d}
Yiming Xie, Chun-Han Yao, Vikram Voleti, Huaizu Jiang, and Varun Jampani.
\newblock Sv4d: Dynamic 3d content generation with multi-frame and multi-view consistency.
\newblock \emph{arXiv preprint arXiv:2407.17470}, 2024{\natexlab{c}}.

\bibitem[Xu et~al.(2024{\natexlab{a}})Xu, Cheng, Gao, Wang, Gao, and Shan]{Xu2024InstantMeshE3}
Jiale Xu, Weihao Cheng, Yiming Gao, Xintao Wang, Shenghua Gao, and Ying Shan.
\newblock {InstantMesh}: Efficient {3D} mesh generation from a single image with sparse-view large reconstruction models.
\newblock \emph{arXiv}, 2024{\natexlab{a}}.

\bibitem[Xu et~al.(2023)Xu, Tan, Luan, Bi, Wang, Li, Shi, Sunkavalli, Wetzstein, Xu, and Zhang]{Xu2023DMV3DDM}
Yinghao Xu, Hao Tan, Fujun Luan, Sai Bi, Peng Wang, Jiahao Li, Zifan Shi, Kalyan Sunkavalli, Gordon Wetzstein, Zexiang Xu, and Kai Zhang.
\newblock {DMV3D}: Denoising multi-view diffusion using {3D} large reconstruction model.
\newblock \emph{arXiv}, 2023.

\bibitem[Xu et~al.(2024{\natexlab{b}})Xu, Shi, Yifan, Chen, Yang, Peng, Shen, and Wetzstein]{Xu2024GRMLG}
Yinghao Xu, Zifan Shi, Wang Yifan, Hansheng Chen, Ceyuan Yang, Sida Peng, Yujun Shen, and Gordon Wetzstein.
\newblock {GRM}: Large gaussian reconstruction model for efficient {3D} reconstruction and generation.
\newblock \emph{arXiv}, 2024{\natexlab{b}}.

\bibitem[Ye et~al.(2024)Ye, Wang, Li, Shi, and Wang]{ye2023consistent1to3}
Jianglong Ye, Peng Wang, Kejie Li, Yichun Shi, and Heng Wang.
\newblock Consistent-1-to-3: Consistent image to 3{D} view synthesis via geometry-aware diffusion models.
\newblock In \emph{3DV}, 2024.

\bibitem[Young(2024)]{xatlas}
Jonathan Young.
\newblock xatlas.
\newblock \url{https://github.com/jpcy/xatlas}, 2024.

\bibitem[yue Li et~al.(2024)yue Li, Long, Liang, Li, Liu, Li, Chi, Qi, Xue, Luo, fei Liu, and Guo]{Li2024MLRMML}
Meng yue Li, Xiaoxiao Long, Yixun Liang, Weiyu Li, Yuan Liu, Peng Li, Xiaowei Chi, Xingqun Qi, Wei Xue, Wenhan Luo, Qi fei Liu, and Yike Guo.
\newblock {M-LRM}: Multi-view large reconstruction model.
\newblock \emph{arXiv}, 2024.

\bibitem[Zaal(2024)]{polyhaven}
Greg Zaal.
\newblock Poly haven, 2024.
\newblock https://polyhaven.com/.

\bibitem[Zhang et~al.(2021)Zhang, Luan, Wang, Bala, and Snavely]{zhang2021}
Kai Zhang, Fujun Luan, Qianqian Wang, Kavita Bala, and Noah Snavely.
\newblock Phy{SG}: Inverse rendering with spherical {G}aussians for physics-based material editing and relighting.
\newblock \emph{CVPR}, 2021.

\bibitem[Zhang et~al.(2024{\natexlab{a}})Zhang, Bi, Tan, Xiangli, Zhao, Sunkavalli, and Xu]{Zhang2024GSLRMLR}
Kai Zhang, Sai Bi, Hao Tan, Yuanbo Xiangli, Nanxuan Zhao, Kalyan Sunkavalli, and Zexiang Xu.
\newblock {GS-LRM}: Large reconstruction model for {3D} gaussian splatting.
\newblock \emph{arXiv}, 2024{\natexlab{a}}.

\bibitem[Zhang et~al.(2018)Zhang, Isola, Efros, Shechtman, and Wang]{zhang2018perceptual}
Richard Zhang, Phillip Isola, Alexei~A Efros, Eli Shechtman, and Oliver Wang.
\newblock The unreasonable effectiveness of deep features as a perceptual metric.
\newblock \emph{CVPR}, 2018.

\bibitem[Zhang et~al.(2024{\natexlab{b}})Zhang, Liu, Xie, Yang, Liu, Yang, Zhang, Kou, Lin, Wang, and Jin]{Zhang2024DreamMatHP}
Yuqing Zhang, Yuan Liu, Zhiyu Xie, Lei Yang, Zhongyuan Liu, Mengzhou Yang, Runze Zhang, Qilong Kou, Cheng Lin, Wenping Wang, and Xiaogang Jin.
\newblock {DreamMat}: High-quality {PBR} material generation with geometry- and light-aware diffusion models.
\newblock \emph{arXiv}, 2024{\natexlab{b}}.

\bibitem[Zhao et~al.(2024)Zhao, Wang, Wang, Zhou, and Zhu]{Zhao2024FlexiDreamerSI}
Ruowen Zhao, Zhengyi Wang, Yikai Wang, Zihan Zhou, and Jun Zhu.
\newblock {FlexiDreamer}: Single image-to-{3D} generation with flexicubes.
\newblock \emph{arXiv}, 2024.

\bibitem[Zheng and Vedaldi(2023)]{zheng2023free3d}
Chuanxia Zheng and Andrea Vedaldi.
\newblock {Free3D}: Consistent novel view synthesis without {3D} representation.
\newblock \emph{arXiv}, 2023.

\bibitem[Zhuang et~al.(2024)Zhuang, Han, Wang, Siarohin, Zou, Vasilkovsky, Shakhrai, Korolev, Tulyakov, and Lee]{Zhuang2024GTRIL}
Peiye Zhuang, Songfang Han, Chaoyang Wang, Aliaksandr Siarohin, Jiaxu Zou, Michael Vasilkovsky, Vladislav Shakhrai, Sergey Korolev, S. Tulyakov, and Hsin-Ying Lee.
\newblock {GTR}: Improving large {3D} reconstruction models through geometry and texture refinement.
\newblock \emph{arXiv}, 2024.

\bibitem[Zou et~al.(2023)Zou, Yu, Guo, Li, Liang, Cao, and Zhang]{Zou2023TriplaneMG}
Zi-Xin Zou, Zhipeng Yu, Yuanchen Guo, Yangguang Li, Ding Liang, Yan-Pei Cao, and Song-Hai Zhang.
\newblock Triplane meets gaussian splatting: Fast and generalizable single-view {3D} reconstruction with transformers.
\newblock \emph{arXiv}, 2023.

\end{thebibliography}
}

\clearpage
\newpage

\setcounter{table}{0}
\setcounter{figure}{0}
\setcounter{section}{0}
\renewcommand{\thetable}{A\arabic{table}}
\renewcommand{\thefigure}{A\arabic{figure}}
\renewcommand\thesection{A\arabic{section}}

\twocolumn[{%
    \renewcommand\twocolumn[1][]{#1}%
    \part*{Supplementary Material}
    \input{fig/triplane_transformer}
}]

\section{Enhanced Transformer}
To reduce the aliasing artifact, we upgrade the transformer backbone to produce triplanes at a resolution of $384 \times 384$. 
However, naively increasing the triplane tokens in TripoSR~\cite{tochilkin2024triposr} is computationally prohibitive due to the quadratic complexity of self-attention. 
Inspired by PointInfinity~\cite{huang2024pointinfinity}, we leverage a two-stream transformer, which has linear complexity w.r.t. the number of tokens. 
As illustrated in \fig{twostreaminterleave}, our architecture consists of two processing streams, the triplane stream and the latent stream. 
The triplane stream consists of the raw triplane tokens to be processed.
In each two-stream unit (gray box in \fig{twostreaminterleave}), the latent stream fetches information from the triplane stream using cross attention, and performs the main computation upon a set of constant-sized latent tokens.
The latent stream then updates the triplane stream with the processed latent tokens.
Our full architecture consists of four such two-stream units.
With this computationally detached design, our transformer is able to produce triplanes at a resolution of $96 \times 96$ with 1024 channels. 
To further increase the resolution and reduce aliasing, we integrated a pixel shuffling operation~\cite{shi2016realtime}, enhancing the triplane resolution to $384 \times 384$ with a feature dimension of 40.

\section{Image metrics}

\begin{table}[ht!]
\centering
\resizebox{0.9\linewidth}{!}{ %
    \begin{tabular}{lcccccc}
           & \multicolumn{3}{c}{GSO} & \multicolumn{3}{c}{OmniObject} \\
           \cmidrule(lr){2-4} \cmidrule{5-7}
        Method & PSNR\textuparrow & SSIM \textuparrow & LPIPS\textdownarrow &PSNR\textuparrow & SSIM \textuparrow & LPIPS\textdownarrow \\
        \midrule
        OpenLRM~\cite{openlrm}               & 15.689 & 0.787 & 0.206 & 13.975 & \secondbest{0.760} & 0.229\\
        TripoSR~\cite{tochilkin2024triposr}  & \secondbest{16.445} & \secondbest{0.789} & \secondbest{0.194} & \secondbest{14.331} & 0.755 & \secondbest{0.224} \\
        LGM~\cite{Tang2024LGMLM} & 14.377 & 0.762 & 0.248 & 12.662 & 0.732 & 0.273
        \\
        CRM~\cite{Wang2024CRMSI} & 15.054 & 0.778 & 0.228 & 13.462 & 0.755 & 0.245 \\
        InstantMesh~\cite{Xu2024InstantMeshE3} & 15.434 & 0.785 & 0.203 & 13.531 & 0.757 & 0.235 \\
        \midrule
        Ours            & \best{21.247} & \best{0.865} & \best{0.124} & \best{20.134} & \best{0.851} & \best{0.132} \\

    \end{tabular}
}
\titlecaption{Comparison on Image Metrics}{} 
\label{tab:comparison_img}
\end{table}

For the image metrics, we follow the pipeline of the shape metric calculation. We further scale the normalized objects back to the actual GT scale and run another finer ICP optimization to adjust the fine scale. 
This transform is then used to render meshes.
This can still result in texture misalignment for highly symmetrical objects, so we treat the image metrics as an auxiliary metric to evaluate the final quality of the asset reconstructions.
Therefore, we only report this metric in the supplements.
\Table{comparison_img} also then supports the improved visual quality during rendering seen in the main paper.


\end{document}